\documentclass[10pt,twocolumn,letterpaper]{article}

\usepackage{sty/cvpr}
\usepackage{times}
\usepackage{epsfig}
\usepackage{graphicx}
\usepackage{amsmath}
\usepackage{amssymb}

\usepackage{url}  %
\usepackage{graphicx, color, xspace}
\usepackage{subcaption}
\usepackage{booktabs}
\usepackage{multirow}
\usepackage{mathrsfs}

\usepackage[pagebackref=true,breaklinks=true,letterpaper=true,colorlinks,bookmarks=false]{hyperref}

\cvprfinalcopy %

\def\cvprPaperID{4450} %

\begin{document}

	\title{EfficientDet: Scalable and Efficient Object Detection}

	\author{
		\begin{tabular}{c c c}
			Mingxing Tan &
			Ruoming Pang &
			Quoc V. Le \\
			\multicolumn{3}{c}{Google Research, Brain Team} \\
			\multicolumn{3}{c}{\{\tt\small tanmingxing, rpang, qvl\}@google.com} \\
		\end{tabular}
	}

	\maketitle
\def\TODO{\textcolor{red}{\emph{TODO: }}}
\newcommand{\TT}[1]{\texttt{#1}}
\newcommand{\BF}[1]{\textbf{#1}}
\newcommand{\IT}[1]{\textit{#1}}

\newcommand\blfootnote[1]{%
	\begingroup
	\renewcommand\thefootnote{}\footnote{#1}%
	\addtocounter{footnote}{-1}%
	\endgroup
}

\newcommand{\km}[1]{{\color{red}[km: #1]}}                                          
\newcommand{\rbg}[1]{{\color{blue}[rbg: #1]}}                                       
\newcommand{\ppd}[1]{{\color{green}[ppd: #1]}}                                      
\newcommand{\bd}[1]{\textbf{#1}}                                                    
\newcommand{\app}{\raise.17ex\hbox{$\scriptstyle\sim$}}                             
\newcommand{\symb}[1]{{\small\texttt{#1}}\xspace}                                   
\newcommand{\mrtwo}[1]{\multirow{2}{*}{#1}}                                         
\def\x{\times}                                                                      
\def\pt{p_\textrm{t}}                                                               
\def\at{\alpha_\textrm{t}}                                                          
\def\xt{x_\textrm{t}}                                                               
\def\CE{\textrm{CE}}                                                                
\def\FL{\textrm{FL}}                                                                
\def\FQ{\textrm{FL}^*}                                                              
\newcommand{\eqnnm}[2]{\begin{equation}\label{eq:#1}#2\end{equation}\ignorespaces}

\newlength\savewidth\newcommand\shline{\noalign{\global\savewidth\arrayrulewidth 
		\global\arrayrulewidth 1pt}\hline\noalign{\global\arrayrulewidth\savewidth}}   
\newcommand{\tablestyle}[2]{\setlength{\tabcolsep}{#1}\renewcommand{\arraystretch}{#2}\centering\footnotesize}
\renewcommand{\dbltopfraction}{1}                                                   
\renewcommand{\bottomfraction}{0}                                                   
\renewcommand{\textfraction}{0}                                                     
\renewcommand{\dblfloatpagefraction}{0.95}                                          
\setcounter{dbltopnumber}{5}

\newcommand{\M}[1]{\mathcal{#1}}
\begin{abstract}

Model efficiency has become increasingly important in computer vision. In this paper, we systematically study neural network architecture design choices for object detection and propose several key optimizations to improve efficiency. First, we propose a weighted bi-directional feature pyramid network (BiFPN), which allows easy and fast multi-scale feature fusion; Second, we propose a compound scaling method that uniformly scales the resolution, depth, and width for all backbone, feature network, and box/class prediction networks at the same time. Based on these optimizations and better backbones, we have developed a new family of object detectors, called EfficientDet, which consistently achieve much better efficiency than prior art across a wide spectrum of resource constraints. In particular, with single-model and single-scale, our EfficientDet-D7 achieves state-of-the-art  \BF{55.1 AP} on COCO \TT{test-dev}  with 77M parameters and 410B FLOPs\footnote{Similar to \cite{resnet16,efficientnet19}, FLOPs denotes number of multiply-adds.},  being \BF{4x -- 9x} smaller and using \BF{13x -- 42x} fewer FLOPs than previous detectors. Code is available at  {\footnotesize \url{ https://github.com/google/automl/tree/master/efficientdet }}.
\end{abstract} 
\section{Introduction}
\label{sec:intro}

Tremendous progresses have been made in recent years towards more accurate object detection; meanwhile, state-of-the-art object detectors also become increasingly more expensive. For example, the latest AmoebaNet-based NAS-FPN detector \cite{odaa19} requires 167M parameters and 3045B FLOPs (30x more than RetinaNet \cite{retinanet17}) to achieve state-of-the-art accuracy. The large model sizes and expensive computation costs deter their deployment in many real-world applications such as robotics and self-driving cars where model size and latency are highly constrained. Given these real-world resource constraints, model efficiency becomes increasingly important for object detection.

\begin{figure}
	\centering
	\includegraphics[width=1.0\columnwidth]{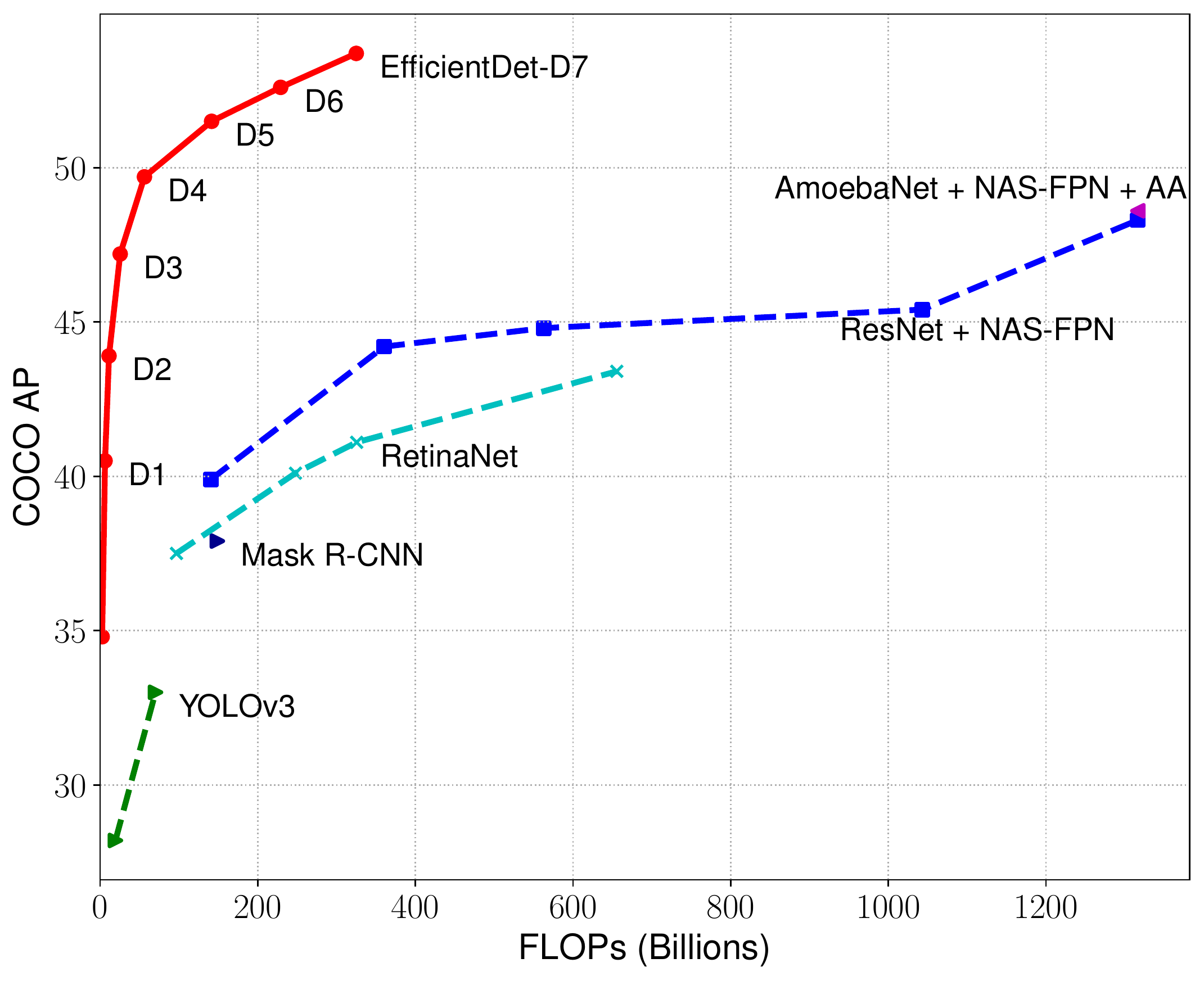}
	\hspace{-47mm}\resizebox{.53\columnwidth}{!}{\tablestyle{2pt}{1}
	\begin{tabular}[b]{l|ll}
		& AP & FLOPs (ratio)\\
		\toprule
		\bf EfficientDet-D0 & \bf  33.8  & \bf 2.5B \\
		YOLOv3 \cite{yolov318} & 33.0 & 71B (28x)   \\
		\hline
		\bf EfficientDet-D1    & \bf 39.6 & \bf 6.1B \\
		RetinaNet \cite{retinanet17}         &  39.2 & 97B (16x) \\
		\hline
		\bf EfficientDet-D7x$^\dagger$                    & \bf 55.1  & \bf 410B \\
		AmoebaNet+ NAS-FPN +AA \cite{odaa19}$^\dagger$      &  50.7      & 3045B (13x)  \\
		\multicolumn{3}{l}{~~$^\dagger$Not plotted. ~~\vspace{11mm} } \\
\end{tabular}}
   \vskip -0.1in
	\caption{\BF{Model FLOPs vs. COCO accuracy --} All numbers are for single-model single-scale. Our EfficientDet achieves new state-of-the-art 55.1\% COCO AP with much fewer parameters and FLOPs than previous detectors. More studies on  different backbones and FPN/NAS-FPN/BiFPN are in Table \ref{tab:backbonefpn} and \ref{tab:bifpncompare}. Complete results are in Table \ref{tab:coco}.
   }

   \vskip -0.2in
	\label{fig:efficientdet-flops}
\end{figure}

There have been many previous works aiming to develop more efficient detector architectures, such as one-stage \cite{ssd16,yolo17,yolov318,retinanet17} and anchor-free detectors \cite{cornetnet18,objectpoints19,fcos19}, or compress existing models  \cite{rethinkprune18,yololite18}.  Although these methods tend to achieve better  efficiency, they usually sacrifice accuracy. Moreover,  most previous works only focus on a specific or a small range of resource requirements, but the variety of real-world applications, from mobile devices to datacenters, often demand different resource constraints.

A natural question is: Is it possible to build a \emph{scalable detection architecture} with both \emph{higher accuracy} and \emph{better efficiency}  across  a wide spectrum of resource constraints (\eg, from 3B to 300B FLOPs)? This paper aims to tackle this problem by systematically studying various design choices of detector architectures. Based on the one-stage detector paradigm, we examine the design choices for backbone, feature fusion, and class/box  network, and identify two main challenges:

\emph{Challenge 1: efficient multi-scale feature fusion -- }  Since  introduced in \cite{fpn17},  FPN has been widely used for multi-scale feature fusion. Recently, PANet \cite{panet18}, NAS-FPN \cite{nasfpn19}, and other studies \cite{deepfpn18,parallelFPN18,m2det19} have developed more network structures for cross-scale feature fusion. While fusing different input features, most previous works simply sum them up without distinction; however, since these different input features are at different resolutions, we observe they usually contribute to the fused output feature unequally. To address this issue, we propose a simple yet highly effective weighted bi-directional feature pyramid network (BiFPN), which introduces learnable weights to learn the importance of different input features, while repeatedly applying top-down and bottom-up multi-scale feature fusion.

\emph{Challenge 2: model scaling  -- } While previous works mainly  rely on bigger backbone networks \cite{retinanet17,fasterrcnn15,yolov318,nasfpn19} or larger input image sizes \cite{maskrcnn17,odaa19} for higher accuracy, we observe that scaling up feature network and  box/class prediction network is also critical when taking into account both accuracy and efficiency. Inspired by recent works \cite{efficientnet19}, we propose a compound scaling method for object detectors, which jointly scales up the resolution/depth/width for all backbone, feature network, box/class prediction network.

Finally, we also observe that the recently introduced EfficientNets \cite{efficientnet19} achieve better  efficiency than previous commonly used backbones. Combining EfficientNet backbones with our propose BiFPN and compound scaling, we have developed a new family of object detectors, named EfficientDet, which consistently achieve better accuracy with  much fewer parameters and FLOPs  than previous object detectors. Figure \ref{fig:efficientdet-flops} and Figure \ref{fig:params-latency} show the performance comparison on COCO dataset \cite{coco14}. Under similar accuracy constraint, our EfficientDet uses 28x fewer FLOPs than YOLOv3 \cite{yolov318}, 30x fewer FLOPs than RetinaNet \cite{retinanet17}, and 19x fewer FLOPs than the recent ResNet based NAS-FPN \cite{nasfpn19}.  In particular, with single-model and single test-time scale, our EfficientDet-D7 achieves state-of-the-art  55.1 AP with 77M parameters and 410B FLOPs,  outperforming previous best detector \cite{odaa19} by 4 AP while being 2.7x smaller and using 7.4x fewer FLOPs. Our EfficientDet is also up to 4x to 11x faster on GPU/CPU than previous detectors.

With simple modifications, we also demonstrate that our single-model single-scale EfficientDet achieves 81.74\% mIOU accuracy with 18B FLOPs on Pascal VOC 2012 semantic segmentation, outperforming DeepLabV3+ \cite{deeplabv3plus14} by 1.7\% better accuracy with 9.8x fewer FLOPs.

\section{Related Work}
\label{sec:related}

\paragraph{One-Stage Detectors: } Existing object detectors are mostly categorized by whether they have a region-of-interest proposal step (two-stage \cite{fastrcnn15,fasterrcnn15,cascade18,maskrcnn17}) or not (one-stage \cite{overfeat14,ssd16,yolo17,retinanet17}).
While two-stage detectors tend to be more flexible and more accurate, one-stage detectors are often considered to be simpler and more efficient by leveraging predefined anchors \cite{speedod17}. Recently,  one-stage detectors have attracted substantial attention  due to th\texttt{}eir efficiency and simplicity \cite{cornetnet18,m2det19,objectpoints19}. In this paper, we mainly follow the one-stage detector design, and we show it is possible to achieve both better efficiency and higher accuracy with optimized network architectures.

\medbreak \noindent \BF{Multi-Scale Feature Representations: } One of the main difficulties in object detection is to effectively represent and process multi-scale features. Earlier detectors often directly perform predictions based on the pyramidal feature hierarchy extracted from backbone networks \cite{multiscaleod16,ssd16,overfeat14}. As one of the pioneering works, feature pyramid network  (FPN) \cite{fpn17} proposes a top-down pathway to combine multi-scale features. Following this idea, PANet \cite{panet18} adds an extra bottom-up path aggregation network on top of FPN; STDL \cite{stdn18} proposes a scale-transfer module to exploit cross-scale features; M2det \cite{m2det19} proposes a U-shape module to fuse multi-scale features, and G-FRNet \cite{gatelabel17} introduces gate units for controlling information flow across features. More recently, NAS-FPN \cite{nasfpn19} leverages neural architecture search to automatically design feature network topology. Although it achieves better performance, NAS-FPN requires thousands of GPU hours during search, and the resulting feature network is irregular and thus difficult to interpret. In this paper, we aim to optimize multi-scale feature fusion with a more intuitive and principled way.

\medbreak \noindent \BF{Model Scaling: }  In order to obtain better accuracy, it is common to scale up a baseline detector by employing bigger backbone networks (\eg, from mobile-size models \cite{mnas19,mobilenetv319} and ResNet \cite{resnet16},  to ResNeXt \cite{resnext17} and AmoebaNet \cite{amoebanets18}),  or increasing input image size (\eg, from 512x512 \cite{retinanet17} to 1536x1536 \cite{odaa19}). Some recent works \cite{nasfpn19,odaa19} show that increasing the channel size and repeating feature networks can also lead to higher accuracy. These scaling methods mostly focus on single or limited scaling dimensions. Recently, \cite{efficientnet19} demonstrates remarkable model efficiency  for image classification by jointly scaling up network width, depth, and resolution. Our proposed compound scaling method for object detection is mostly inspired by \cite{efficientnet19}.

\section{BiFPN}
\label{sec:bifpn}

\begin{figure*}
	    \centering
        \includegraphics[width=0.99\linewidth]{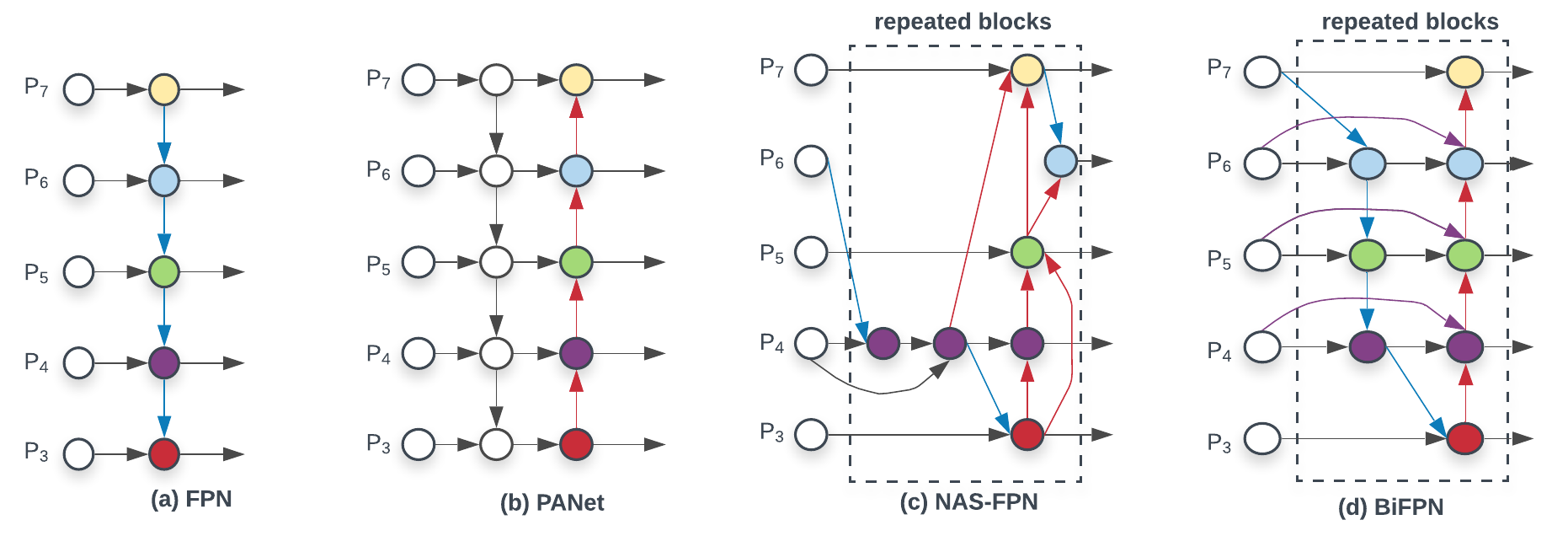}
        \caption{\BF{Feature network design --}
        	(a) FPN \cite{fpn17} introduces a top-down pathway to fuse multi-scale features from level 3 to 7 ($P_3$ - $P_7$); (b) PANet \cite{panet18} adds an additional bottom-up pathway on top of FPN; (c)  NAS-FPN \cite{nasfpn19} use neural architecture search to find an irregular feature network topology and then repeatedly apply the same block; (d) is our BiFPN with better accuracy and efficiency trade-offs.
        }
       \vskip -0.1in
        \label{fig:bifpn}                                                  
\end{figure*} 
 
In this section, we first formulate the multi-scale feature fusion problem, and then introduce the main ideas for our proposed BiFPN: efficient bidirectional cross-scale connections and weighted feature fusion.

\subsection{Problem Formulation}
\label{subsec:problem}

Multi-scale feature fusion aims to aggregate features at different resolutions. Formally, given a list of multi-scale features  $\vec{P}^{in} = (P^{in}_{l_1}, P^{in}_{l_2}, ...)$, where $P^{in}_{l_i}$ represents the feature at level $l_i$, our goal is to find a transformation $f$ that can effectively aggregate different features and output a list of new features: $\vec{P}^{out} = f(\vec{P}^{in})$. As a concrete example, Figure \ref{fig:bifpn}(a) shows the conventional top-down FPN \cite{fpn17}. It takes level 3-7 input features $\vec{P}^{in} = (P^{in}_{3}, ... P^{in}_{7})$, where $P^{in}_{i}$ represents a feature level with resolution of $1/2^i$ of the input images. For instance, if input resolution is 640x640, then $P^{in}_{3}$ represents feature level 3 ($640/2^3=80$) with resolution 80x80, while  $P^{in}_{7}$ represents feature level 7 with resolution 5x5.  The conventional FPN aggregates multi-scale features in a top-down manner:

\begin{align*}
P^{out}_7 &= Conv(P^{in}_7)  \\
P^{out}_6 &= Conv(P^{in}_6 + Resize(P^{out}_7)) \\
... \\
P^{out}_3 &= Conv(P^{in}_3 + Resize(P^{out}_4))
\end{align*}

\noindent where $Resize$  is usually a upsampling or downsampling op for resolution matching, and
$Conv$ is usually a convolutional op for feature processing.

\subsection{Cross-Scale Connections}
\label{subsec:connection}

Conventional top-down FPN  is inherently limited by the one-way information flow. To address this issue, PANet \cite{panet18} adds an extra bottom-up path aggregation network, as shown in Figure \ref{fig:bifpn}(b). Cross-scale connections are further studied in \cite{deepfpn18,parallelFPN18, m2det19}. Recently, NAS-FPN \cite{nasfpn19} employs neural architecture search to search for better cross-scale feature network topology, but it requires thousands of GPU hours during search and the found network is irregular and difficult to interpret or modify, as shown in Figure \ref{fig:bifpn}(c).

By studying the performance and efficiency of these three networks (Table \ref{tab:bifpncompare}), we observe that PANet achieves better accuracy than FPN and NAS-FPN, but with the cost of more parameters and  computations. To improve model efficiency, this paper proposes several optimizations for cross-scale connections: First, we remove those nodes that only have one input edge. Our intuition is simple: if a node has only one input edge with no feature  fusion, then it will have less contribution to feature network that aims at fusing different features. This leads to a simplified bi-directional network;  Second, we add an extra edge from the original input to output node if they are at the same level, in order to fuse more features without adding much cost; Third, unlike PANet \cite{panet18} that only has one top-down and one bottom-up path, we treat each bidirectional (top-down \& bottom-up) path as one feature network layer, and repeat the same layer multiple times to enable more high-level feature fusion. Section \ref{subsec:scaling} will discuss how to determine the number of layers for different resource constraints using a compound scaling method.  With these optimizations, we name the new feature network as bidirectional feature pyramid network (BiFPN), as shown in Figure \ref{fig:bifpn} and  \ref{fig:arch}.

\subsection{Weighted Feature Fusion}
\label{subsec:attention}

When fusing features with different resolutions, a common way is to first resize them to the same resolution and then sum them up.  Pyramid attention network \cite{pan18} introduces global self-attention upsampling to recover pixel localization, which is further studied in \cite{nasfpn19}. All previous methods treat all input features equally without distinction. However, we observe that since different input features are at different resolutions, they usually contribute to the output feature \emph{unequally}. To address this issue, we propose to add an additional weight for each input, and let the network to learn the importance of each input feature. Based on this idea, we consider three weighted fusion approaches:

\medbreak
\noindent \BF{Unbounded fusion:} $O = \sum_{i} w_i \cdot I_i $, where $w_i$ is a learnable weight that can be a scalar (per-feature), a vector (per-channel), or a multi-dimensional tensor (per-pixel). We find a scale can achieve comparable accuracy to other approaches with minimal computational costs. However, since the scalar weight is unbounded, it could potentially cause training instability. Therefore, we resort to weight normalization to bound the value range of each weight.

\noindent \BF{Softmax-based fusion:} $O = \sum_{i} \cfrac{e^{w_i}}{\sum_{j} e^{w_j}} \cdot I_i $. An intuitive idea is to apply softmax to each weight, such that all weights are normalized to be a probability with value range from 0 to 1, representing the importance of each input. However, as shown in our ablation study in section \ref{subsec:ablation-att}, the extra softmax leads to significant slowdown on GPU hardware. To minimize the extra latency cost, we further propose a fast fusion approach.

\noindent \BF{Fast normalized fusion:} $O = \sum_{i} \cfrac{w_i}{\epsilon + \sum_{j} w_j} \cdot I_i$, where $w_i \ge 0$ is ensured by applying a Relu after each $w_i$, and $\epsilon = 0.0001$ is a  small value to avoid numerical instability. Similarly, the value of each normalized weight also falls between 0 and 1, but since there is no softmax operation here, it is much more efficient. Our ablation study shows this fast fusion approach has very similar  learning behavior and accuracy as the softmax-based fusion, but runs up to 30\% faster on GPUs (Table \ref{tab:attnacc}).

\medbreak
Our final BiFPN integrates both the bidirectional cross-scale connections and the fast normalized fusion. As a concrete example, here we describe the two fused features at level 6 for BiFPN shown in Figure \ref{fig:bifpn}(d):

\begin{align*} 
P^{td}_6 &= Conv\left(\frac{w_1\cdot P^{in}_6  +     w_2 \cdot Resize(P^{in}_7)}{w_1 + w_2 + \epsilon}\right) \\
P^{out}_6 &= Conv\left(\frac{w'_1\cdot P^{in}_6  +  w'_2 \cdot P^{td}_6 + w'_3 \cdot Resize(P^{out}_5)}{w'_1 + w'_2 + w'_3 + \epsilon}\right)
\end{align*}

\noindent where $P^{td}_6$ is the intermediate feature at level 6 on the top-down pathway, and $P^{out}_6$ is the output feature at level 6 on the bottom-up pathway. All other features are constructed in a similar manner. Notably, to further improve the efficiency, we use depthwise separable convolution \cite{xception17,sepconv14} for feature fusion,  and add batch normalization and activation after each convolution.
\section{EfficientDet}
\label{sec:network}

\begin{figure*}
	    \centering
        \includegraphics[width=0.99\linewidth]{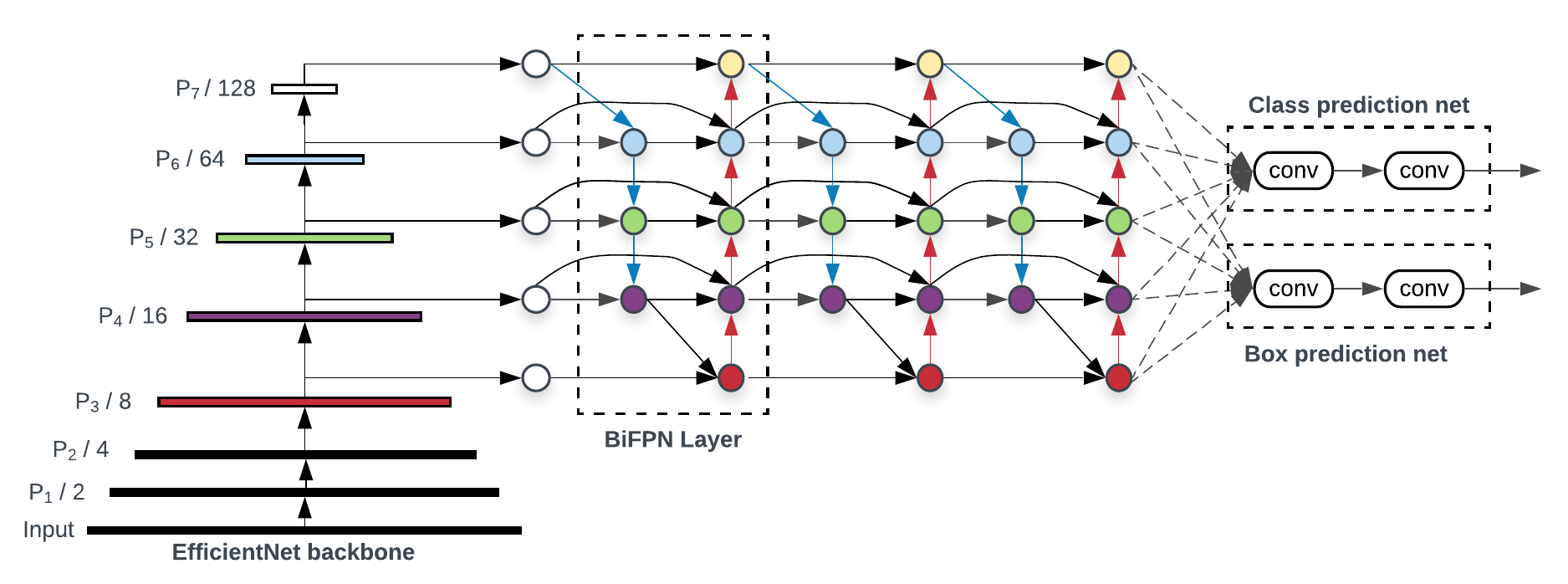}
        \vskip -0.1in
        \caption{\BF{EfficientDet architecture --} It employs EfficientNet \cite{efficientnet19} as the backbone network, BiFPN as the feature network, and shared class/box prediction network. Both BiFPN layers and class/box net layers are repeated multiple times based on different resource constraints as shown in Table \ref{tab:scaleconfigs}.}
        \label{fig:arch}                                            
\end{figure*}

Based on our BiFPN, we have developed a new family of detection models named EfficientDet. In this section, we will discuss the network architecture and a new compound scaling method for EfficientDet.

\subsection{EfficientDet Architecture}
\label{subsec:overview}

Figure \ref{fig:arch} shows the overall architecture of EfficientDet, which largely follows the one-stage detectors paradigm \cite{ssd16,yolo17,fpn17,retinanet17}. We employ ImageNet-pretrained EfficientNets as the backbone network. Our proposed BiFPN serves as the feature network, which takes level 3-7 features \{$P_3, P_4, P_5, P_6, P_7$\} from the backbone network and repeatedly applies top-down and bottom-up bidirectional feature fusion. These fused features are fed to a class and box  network to produce object class  and bounding box predictions respectively. Similar to \cite{retinanet17}, the class and box network weights are shared across all levels of features.

\subsection{Compound Scaling}
\label{subsec:scaling}

Aiming at optimizing both accuracy and efficiency, we would like to develop a family of models  that can meet a wide spectrum of resource constraints. A key challenge here is how to scale up a baseline EfficientDet model.

Previous works mostly scale up a baseline detector by employing bigger backbone networks (\eg, ResNeXt \cite{resnext17} or AmoebaNet \cite{amoebanets18}), using larger input images, or stacking more FPN layers \cite{nasfpn19}. These methods are usually ineffective since they only focus on a single or  limited scaling dimensions.
Recent work \cite{efficientnet19} shows remarkable performance on image classification by jointly scaling up all dimensions of network width, depth, and input resolution. Inspired by these works \cite{nasfpn19,efficientnet19}, we propose a new compound scaling method for object detection, which uses a simple compound coefficient $\phi$ to jointly scale up all dimensions of backbone , BiFPN, class/box network, and resolution. Unlike \cite{efficientnet19}, object detectors have much more scaling dimensions than image classification models, so grid search for all dimensions is prohibitive expensive. Therefore, we use a heuristic-based scaling approach, but still follow the main idea of jointly scaling up all dimensions.

\paragraph{Backbone network --} we  reuse the same width/depth scaling coefficients of  EfficientNet-B0 to B6 \cite{efficientnet19} such that we can easily reuse their ImageNet-pretrained checkpoints.

\paragraph{BiFPN network --} we linearly increase BiFPN depth $D_{bifpn}$ (\#layers) since depth needs to be rounded to small integers. For BiFPN width $W_{bifpn}$  (\#channels), exponentially grow BiFPN width $W_{bifpn}$ (\#channels) as similar to \cite{efficientnet19}. Specifically, we perform a grid search on a list of values \{1.2, 1.25, 1.3, 1.35, 1.4, 1.45\}, and pick the best value 1.35 as the BiFPN width scaling factor.  Formally, BiFPN width and depth are scaled with the following equation:
\begin{equation} \label{eq:scale-bifpn}
W_{bifpn} = 64 \cdot \left(1.35^{\phi}\right), \quad \quad
D_{bifpn} = 3 + \phi
\end{equation}

\paragraph{Box/class  prediction network  -- } we fix their width to be always the same as BiFPN (i.e., $W_{pred} = W_{bifpn}$), but linearly increase the depth (\#layers) using equation:
\begin{equation} \label{eq:scale-pred}
D_{box} = D_{class} = 3 + \lfloor \phi / 3 \rfloor
\end{equation}

\paragraph{Input image resolution -- } Since feature level 3-7 are used in BiFPN, the input resolution must be dividable by $2^7=128$, so we linearly increase resolutions using equation:
\begin{equation} \label{eq:scale-resolution}
R_{input} = 512 + \phi \cdot 128
\end{equation}

\noindent Following Equations \ref{eq:scale-bifpn},\ref{eq:scale-pred},\ref{eq:scale-resolution} with different $\phi$, we have developed EfficientDet-D0  ($\phi=0$) to D7 ($\phi=7$) as shown in Table \ref{tab:scaleconfigs}, where D7 and D7x have the same BiFPN and head, but D7 uses higher resolution and D7x uses larger backbone network and one more feature level (from $P_3$ to $P_8$). Notably, our compound scaling is heuristic-based and might  not be optimal, but we will show that this simple scaling method can significantly improve efficiency than other single-dimension scaling methods in Figure \ref{fig:scale-flops}.

\begin{table}
	\centering
	\resizebox{0.99\columnwidth}{!}{
		\begin{tabular}{l|cccccc}
			\toprule[0.15em]
			\multirow{2}{*}{} &   Input &  Backbone & \multicolumn{2}{c}{BiFPN}  & Box/class \\
			&  size  &  Network & \#channels  & \#layers  & \#layers  \\
			&  $R_{input}$  &   & $W_{bifpn}$  & $D_{bifpn}$  & $D_{class}$  \\
			\midrule[0.1em]
			D0 ($\phi=0$)	&  512 & B0	&	64	&	3	&	3  \\
			D1 ($\phi=1$)	&  640 & B1	&	88	&	4	&	3  \\
			D2 ($\phi=2$)	&  768 & B2	&  112	&	5	&	3  \\
			D3 ($\phi=3$)	&  896 & B3	&  160	&	6	&	4  \\
			D4 ($\phi=4$)	&  1024 & B4	&	224	&	7	&	4  \\
			D5 ($\phi=5$)	&  1280 & B5	&	288	&	7	&	4  \\
			D6 ($\phi=6$)	&  1280 & B6	&	384	&	8	&	5  \\
			D7 ($\phi=7$)	&  1536  & B6	&	384	&	8	&	5  \\
			D7x	&  1536 & B7	&	384	&	8	&	5  \\
		\bottomrule
		\end{tabular}
	}
	\caption{
	\textbf{Scaling configs for EfficientDet D0-D6 -- }  $\phi$ is the compound coefficient that controls all other scaling dimensions; \emph{BiFPN, box/class net, and input size} are scaled up using equation \ref{eq:scale-bifpn}, \ref{eq:scale-pred}, \ref{eq:scale-resolution} respectively.
   }
	\label{tab:scaleconfigs}
\end{table}
\newcommand\sd[1]{{\scriptsize $\pm$#1}}

\begin{table*}

    \centering
    \resizebox{1.0\textwidth}{!}{
        \begin{tabular}{l||ccc|c||rcrc||cc}
        \toprule [0.15em]
         &  \multicolumn{3}{c|}{\TT{test-dev}}  & \TT{val} & \multicolumn{4}{c||}{} & \multicolumn{2}{c}{Latency (ms) } \\
        Model &  $AP$ & $AP_{50}$ &  $AP_{75}$ &  $AP$ & Params &  Ratio &FLOPs & Ratio & TitianV  & V100 \\
        \toprule
        \bf EfficientDet-D0  (512)   & \bf 34.6  & \bf 53.0      & \bf 37.1    &  \bf 34.3 & \bf 3.9M  &  \bf 1x & \bf 2.5B  & \bf 1x  & \bf 12      & \bf 10.2 \\
		YOLOv3 \cite{yolov318}	&	33.0	&	57.9	&	34.4	&	- &  - & - &	71B	&	28x  &	-	&	-	 \\

        \midrule
        \bf EfficientDet-D1  (640)  	& \bf 40.5	 & \bf 59.1		& \bf 43.7 	& \bf 40.2 & \bf 6.6M  &  \bf 1x & \bf 6.1B  & \bf 1x  & \bf 16      & \bf 13.5 \\
        RetinaNet-R50 (640)	\cite{retinanet17} &	39.2	& 58.0 & 42.3 & 39.2 &	34M	& 6.7x		&	97B	&	16x	&	25 	&	- \\
		RetinaNet-R101 (640)\cite{retinanet17} &	39.9	&	58.5 & 43.0 & 39.8 & 53M	& 8.0x		&	127B	&	21x	&	32	&	- \\

        \midrule
        \bf EfficientDet-D2  (768)   	& \bf 43.9	 & \bf 62.7		& \bf 47.6	& \bf 43.5 & \bf 8.1M  &  \bf 1x & \bf 11B  & \bf 1x  & \bf 23      & \bf 17.7 \\
		Detectron2 Mask R-CNN R101-FPN \cite{detectron2} & - &  - & - & 42.9 & 63M & 7.7x & 164B &  15x & - & 56$^\ddag$ \\
		Detectron2 Mask R-CNN X101-FPN \cite{detectron2} & - &   - &  - &  44.3 &  107M &  13x &  277B &  25x & - & 103$^\ddag$ \\
        \midrule

        \bf EfficientDet-D3  (896)  	& \bf 47.2	 & \bf 65.9		& \bf 51.2	& \bf 46.8 & \bf 12M  &  \bf 1x & \bf  25B  & \bf 1x  & \bf 37      & \bf 29.0 \\
		ResNet-50 + NAS-FPN (1024)	\cite{nasfpn19} & 44.2	& - & - & -	&	60M &	5.1x	&	360B 	&	15x	&	64 	&	-	\\
		ResNet-50 + NAS-FPN  (1280)	\cite{nasfpn19} & 44.8	& - & - & -	&	60M &	5.1x	&	563B	&	23x	&	99 	&	-	\\
		ResNet-50 + NAS-FPN (1280@384)\cite{nasfpn19} &	45.4	& - & - & -		&	104M	&	8.7x	&	 1043B	&	42x	&	150 	&	-	\\

        \midrule
        \bf EfficientDet-D4  (1024)   & \bf 49.7	& \bf 68.4		& \bf 53.9 	& \bf 49.3 & \bf 21M  &  \bf 1x & \bf  55B  & \bf 1x  & \bf 65      & \bf 42.8 \\
		AmoebaNet+ NAS-FPN +AA(1280)\cite{odaa19} &	-	& - & - & 48.6	 &	185M	&	8.8x	& 1317B		&	24x	&	246 	&	-	\\

        \midrule
        \bf EfficientDet-D5  (1280)   & \bf 51.5	& \bf 70.5		& \bf 56.1 	& \bf 51.3 & \bf 34M  &  \bf 1x & \bf 135B  & \bf 1x  & \bf 128      & \bf 72.5 \\
		Detectron2 Mask R-CNN X152 \cite{detectron2} & - &   - &  - &  50.2 &  -  &  - &  - &  - & - & 234$^\ddag$ \\
        \midrule
        \bf EfficientDet-D6  (1280)   & \bf 52.6	& \bf 71.5		& \bf 57.2 	& \bf 52.2 & \bf 52M  &  \bf 1x  & \bf 226B  & \bf 1x  & \bf 169      & \bf 92.8 \\
       AmoebaNet+ NAS-FPN +AA(1536)\cite{odaa19}	&	- & - & - & 50.7	&	209M	&	4.0x	& 3045B		&	13x		&	489	&	-		\\
        \midrule
        \bf EfficientDet-D7  (1536)   & \bf 53.7	& \bf 72.4		& \bf 58.4	& \bf 53.4 & \bf 52M  &  \bf   & \bf 325B  &   & \bf 232     & \bf 122 \\
        \bf EfficientDet-D7x  (1536)   & \bf 55.1	& \bf 74.3		& \bf 59.9	& \bf 54.4 & \bf 77M  &  \bf   & \bf 410B  &   & \bf285    & \bf 153 \\
		\bottomrule
        \multicolumn{11}{l}{~We omit ensemble and test-time multi-scale results \cite{megdet18,rethinking19}. RetinaNet APs are reproduced with our trainer and others are from papers.} \\
        \multicolumn{11}{l}{~$^\ddag$Latency numbers with $^\ddag$ are  from detectron2, and others are measured on the same machine (TensorFlow2.1 + CUDA10.1, no TensorRT). }
        \end{tabular}
    }

    \caption{
        \textbf{EfficientDet performance on COCO} \cite{coco14} -- Results are for single-model single-scale.  \TT{test-dev} is the COCO test set and \TT{val} is the validation set. \TT{Params} and \TT{FLOPs} denote the  number of parameters and multiply-adds. \TT{Latency} is for inference  with batch size 1. \TT{AA} denotes auto-augmentation \cite{odaa19}. We group models together if they have similar accuracy, and compare their model size, FLOPs, and latency in each group.
       }

    \label{tab:coco}
\end{table*}

\section{Experiments}
\label{sec:results}

\subsection{EfficientDet for Object Detection}

We evaluate EfficientDet on COCO 2017 detection datasets \cite{coco14} with 118K training images. Each model is trained using SGD optimizer with momentum 0.9 and weight decay 4e-5. Learning rate is linearly increased from 0 to 0.16 in the first  training epoch and then annealed down using cosine decay rule. Synchronized batch norm is added after every convolution with batch norm decay 0.99 and epsilon 1e-3. Same as the \cite{efficientnet19}, we use SiLU (Swish-1) activation \cite{swishsil18,gelu16,swish18} and exponential moving average with decay 0.9998. We also employ commonly-used focal loss \cite{retinanet17} with $\alpha = 0.25$ and $\gamma = 1.5$, and aspect ratio \{1/2, 1, 2\}.  During training, we apply horizontal flipping and scale jittering [0.1, 2.0], which randomly rsizes images between 0.1x and 2.0x of the original size before cropping. We apply soft-NMS \cite{softnms17} for eval. For D0-D6, each model is  trained for 300 epochs with total batch  size 128 on 32 TPUv3 cores, but to push the envelope, we train D7/D7x for 600 epochs on 128 TPUv3 cores.

Table \ref{tab:coco} compares EfficientDet with other object detectors, under the single-model single-scale settings with no test-time augmentation. We report accuracy for both \TT{test-dev}  (20K test images with no public ground-truth) and \TT{val} with 5K validation images. Notably, model performance depends on both network architecture and trainning settings (see appendix), but for simplicity, we only reproduce RetinaNet using our trainers and refer other models from their papers.
In general, our EfficientDet achieves better efficiency than previous detectors, being \BF{4x -- 9x} smaller and using \BF{13x - 42x} less FLOPs across a wide range of accuracy or resource constraints.
On relatively low-accuracy regime, our EfficientDet-D0 achieves similar accuracy as YOLOv3 with 28x fewer FLOPs. Compared to RetinaNet \cite{retinanet17} and Mask-RCNN \cite{maskrcnn17}, our EfficientDet achieves similar accuracy with up to 8x fewer parameters and 21x fewer FLOPs. On high-accuracy regime, our EfficientDet also consistently outperforms recent object detectors \cite{nasfpn19,odaa19} with much fewer parameters and FLOPs. In particular, our single-model single-scale EfficientDet-D7x achieves a new state-of-the-art \BF{55.1} AP on \TT{test-dev}, outperforming prior art by a large margin in both accuracy (+4 AP) and efficiency (7x fewer FLOPs).

\begin{figure*}
    \centering
    \begin{subfigure}[t]{0.33\textwidth}
        \centering
        \includegraphics[width=1.0\columnwidth]{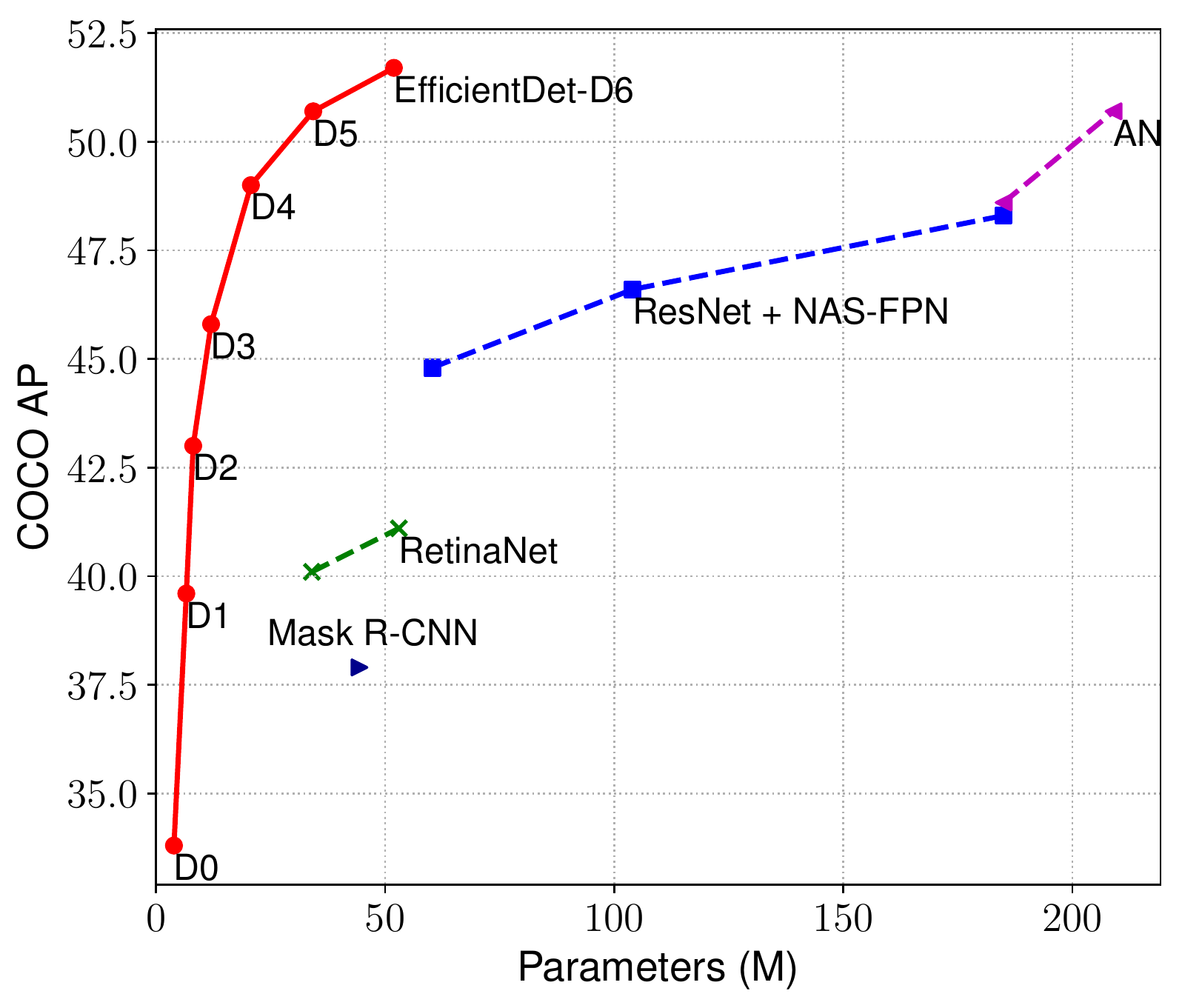}
		\hspace{-35mm}\resizebox{.57\columnwidth}{!}{\tablestyle{2pt}{1}
			\begin{tabular}[b]{l|cc}
				& Params & Ratio\\
				\toprule
				EfficientDet-D1                      & 7M  &  \\
				RetinaNet \cite{retinanet17}         &  53M & \bf 8.0x \\
				\hline
				EfficientDet-D3                      &  12M &  \\
				ResNet + NASFPN \cite{nasfpn19}             &  104M & \bf 8.7x \\
				\hline
				EfficientDet-D6                     & 52M &  \\
				AmoebaNet + NAS-FPN \cite{odaa19}       &  209M      & \bf 4.0x  \\
				\multicolumn{3}{l}{~~\vspace{9mm} }
			\end{tabular}}
			   \vskip -0.03in
        \caption{Model Size}
    \end{subfigure}
    \begin{subfigure}[t]{0.33\textwidth}
        \centering
		\includegraphics[width=1.0\columnwidth]{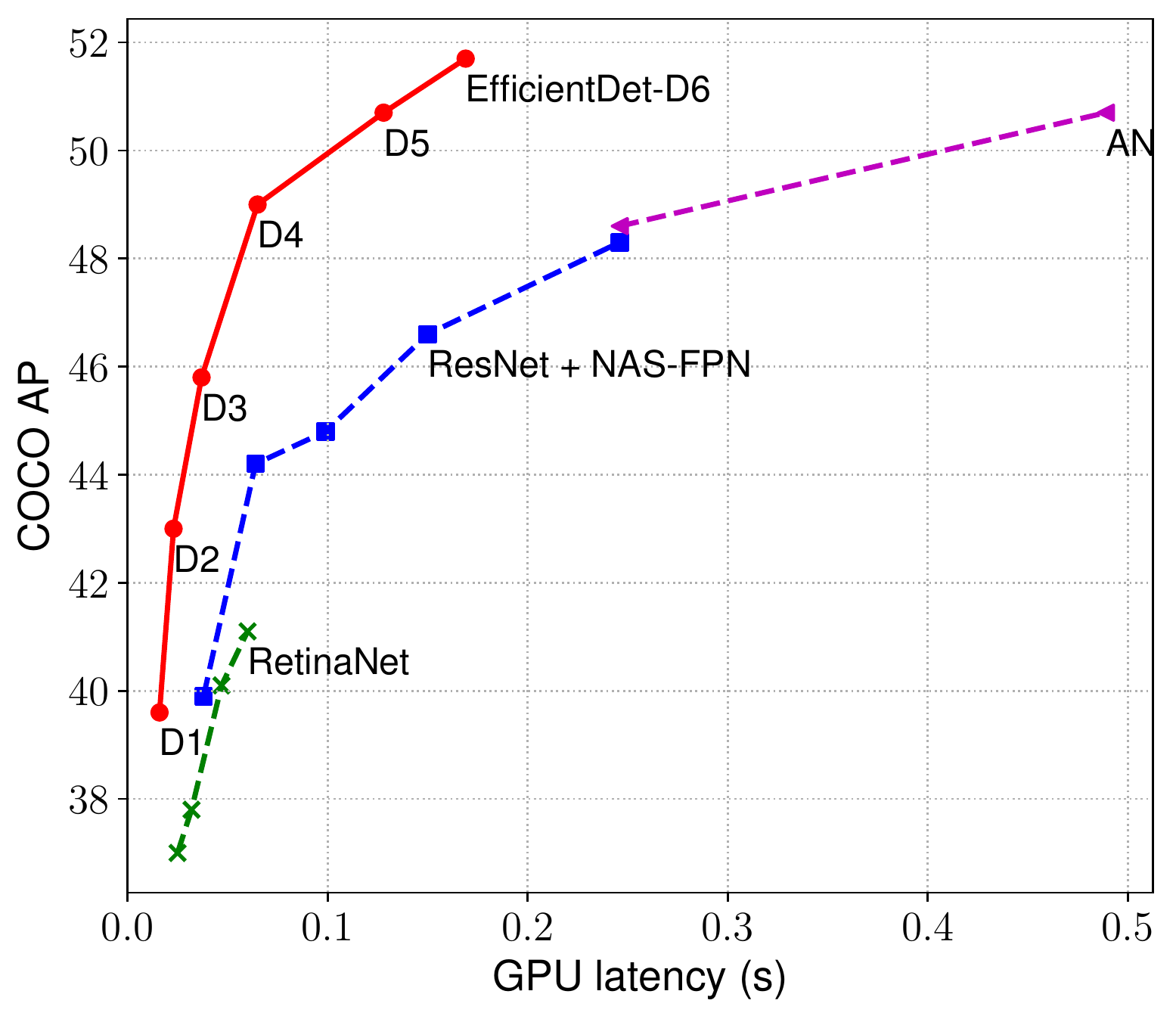}
		\hspace{-36mm}\resizebox{.58\columnwidth}{!}{\tablestyle{2pt}{1}
			\begin{tabular}[b]{l|cc}
				& LAT & Ratio \\
				\toprule
				EfficientDet-D1                      &  16ms  &  \\
				RetinaNet \cite{retinanet17}         &  32ms & \bf 2.0x \\
				\hline
				EfficientDet-D3                      &  37ms &  \\
				ResNet + NASFPN \cite{nasfpn19}             &  150ms & \bf 4.1x \\
				\hline
				EfficientDet-D6                     &  169ms &  \\
				AmoebaNet + NAS-FPN \cite{odaa19}       &  489ms     & \bf 2.9x  \\
				\multicolumn{3}{l}{~~\vspace{9.5mm} }
			\end{tabular}}
			   \vskip -0.03in
		\caption{GPU Latency}
    \end{subfigure}
    \begin{subfigure}[t]{0.33\textwidth}
		\centering
		\includegraphics[width=1.0\columnwidth]{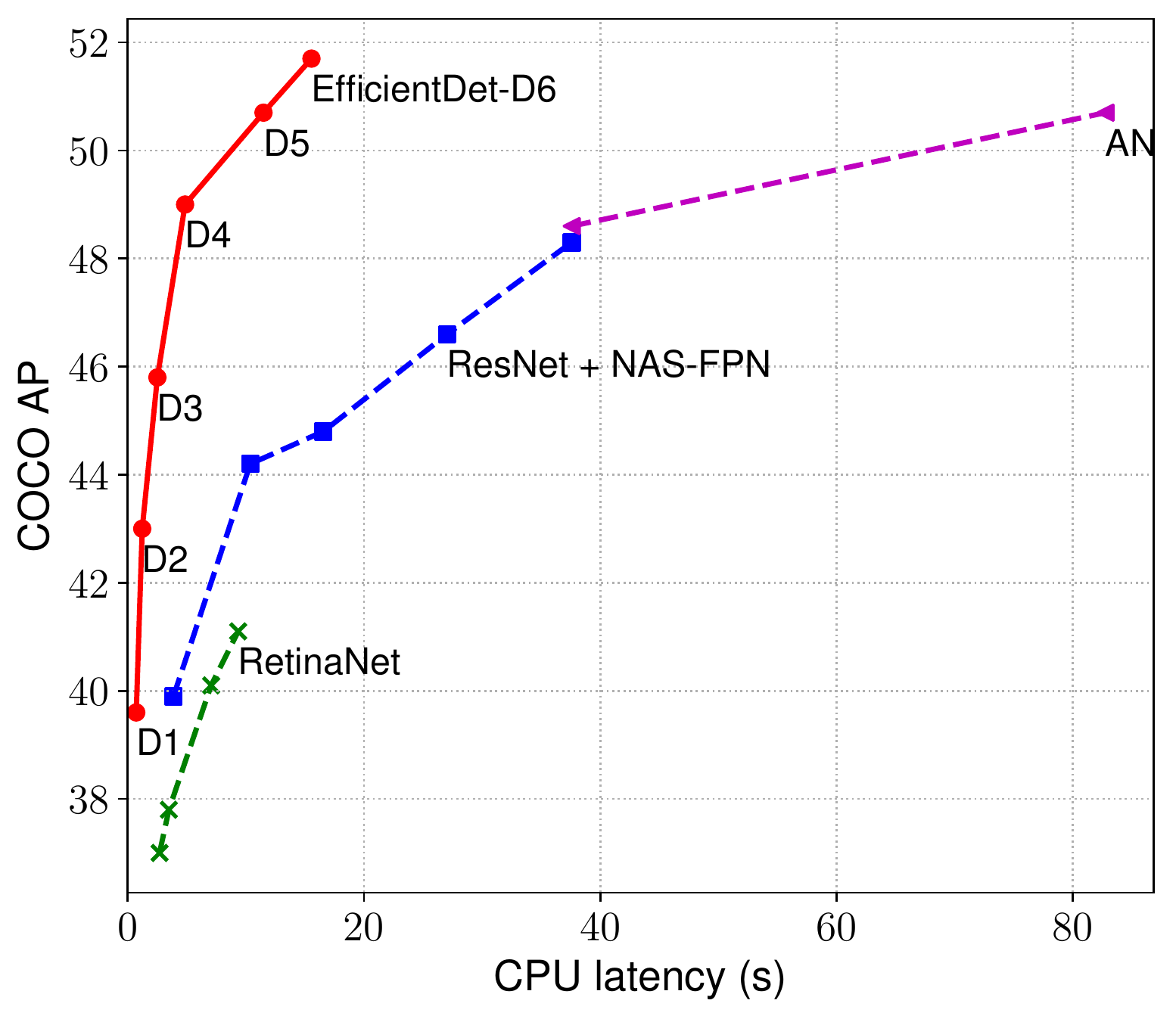}
		\hspace{-34mm}\resizebox{.55\columnwidth}{!}{\tablestyle{2pt}{1}
		\begin{tabular}[b]{l|cc}
			& LAT & Ratio \\
			\toprule
			EfficientDet-D1                      &  0.74s  &  \\
			RetinaNet \cite{retinanet17}         &  3.6s & \bf 4.9x \\
			\hline
			EfficientDet-D3                      &  2.5s &  \\
			ResNet + NASFPN \cite{nasfpn19}             &  27s & \bf 11x \\
			\hline
			EfficientDet-D6    &  16s &  \\
			AmoebaNet + NAS-FPN \cite{odaa19}    &  83s     & \bf 5.2x  \\
				\multicolumn{3}{l}{~~\vspace{9.5mm} }
		\end{tabular}}
			   \vskip -0.03in
		\caption{CPU Latency}
	\end{subfigure}
	\vskip -0.1in
    \caption{
    	\BF{Model size and inference latency comparison}  -- Latency is measured with batch size 1 on the same machine equipped with a Titan V GPU and Xeon CPU. \TT{AN} denotes AmoebaNet + NAS-FPN trained with auto-augmentation \cite{odaa19}. Our EfficientDet models are 4x - 9x smaller, 2x - 4x faster on GPU, and 5x - 11x faster on CPU than other  detectors.
    }
     \label{fig:params-latency}
\end{figure*}
 
In addition, we have also compared the  inference latency on Titan-V FP32 , V100 GPU FP16, and single-thread  CPU. Notably, our V100 latency is end-to-end including preprocessing and NMS postprocessing.
Figure \ref{fig:params-latency} illustrates the comparison on model size and GPU/CPU latency.  For fair comparison, these figures only include results that are measured on the same machine with the same settings. Compared to previous detectors, EfficientDet models are up to 4.1x faster on GPU and 10.8x faster on CPU, suggesting they are also efficient on real-world hardware.

\subsection{EfficientDet for Semantic Segmentation}

While our EfficientDet models are mainly designed for object detection, we are also interested in their performance on other tasks such as semantic segmentation. Following \cite{panopticfpn19}, we modify our EfficientDet model to keep feature level $\{P2, P3, ..., P7\}$ in BiFPN, but only use $P2$ for the final per-pixel classification. For simplicity, here we only evaluate a EfficientDet-D4 based model, which uses a ImageNet pretrained EfficientNet-B4 backbone (similar size to ResNet-50). We set the channel size to 128 for BiFPN and 256 for classification head. Both BiFPN and classification head are repeated by 3 times.

\begin{table}                                            
	\centering   	
	\resizebox{0.95\columnwidth}{!}{ 
		\begin{tabular}{l|ccc}                
			\toprule[0.15em]      
			Model      &    mIOU  & Params &  FLOPs \\ 
			\midrule[0.1em]                  
			DeepLabV3+ (ResNet-101) \cite{deeplabv3plus14}   &  79.35\% & - & 298B  \\
			DeepLabV3+ (Xception) \cite{deeplabv3plus14}   &  80.02\% & - & 177B  \\
			\bf Our EfficientDet$^\dagger$ & \bf 81.74\%  & \bf 17M & \bf 18B    \\
			\bottomrule[0.15em]      
			\multicolumn{4}{l}{~$^\dagger$A modified version of EfficientDet-D4. }                                                     
		\end{tabular}                                                                 
	}
	\caption{                                                                         
	\textbf{Performance comparison on Pascal VOC semantic segmentation}.
   }
   \vskip -0.1in
	\label{tab:seg}
\end{table} 
 
Table \ref{tab:seg} shows the comparison between our models and previous DeepLabV3+ \cite{deeplabv3plus14} on Pascal VOC 2012 \cite{pascalvoc12}. Notably, we exclude those results with ensemble,  test-time augmentation, or COCO pretraining. Under the same single-model single-scale settings, our model achieves 1.7\% better accuracy with 9.8x fewer FLOPs than the prior art of DeepLabV3+ \cite{deeplabv3plus14}. These results suggest that EfficientDet is also quite promising for semantic segmentation.

\section{Ablation Study}
\label{sec:ablation}

\begin{figure*}
	    \centering
        \begin{subfigure}[t]{0.315\textwidth}
        	\includegraphics[width=1.0\linewidth]{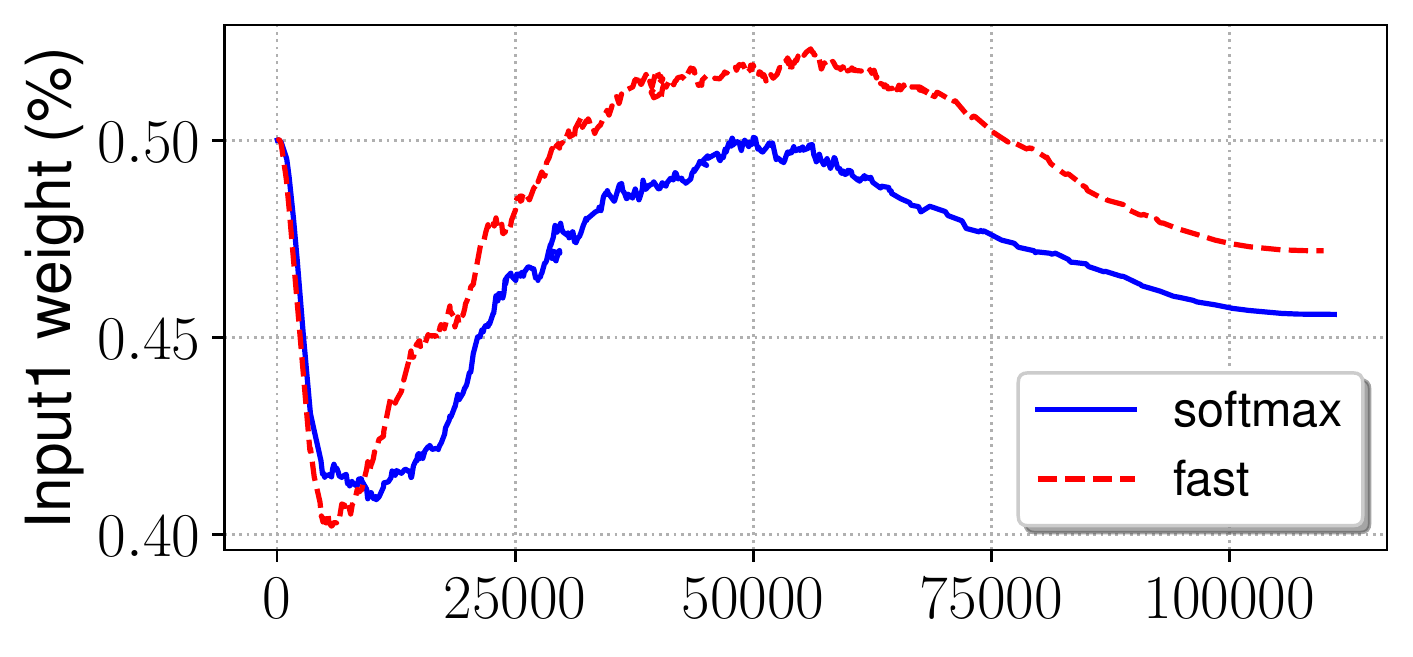}
        	\vskip -0.1in
            \caption{Example Node 1}
	    \end{subfigure}
        \hfill
        \begin{subfigure}[t]{0.31\textwidth}
			\includegraphics[width=1.0\linewidth]{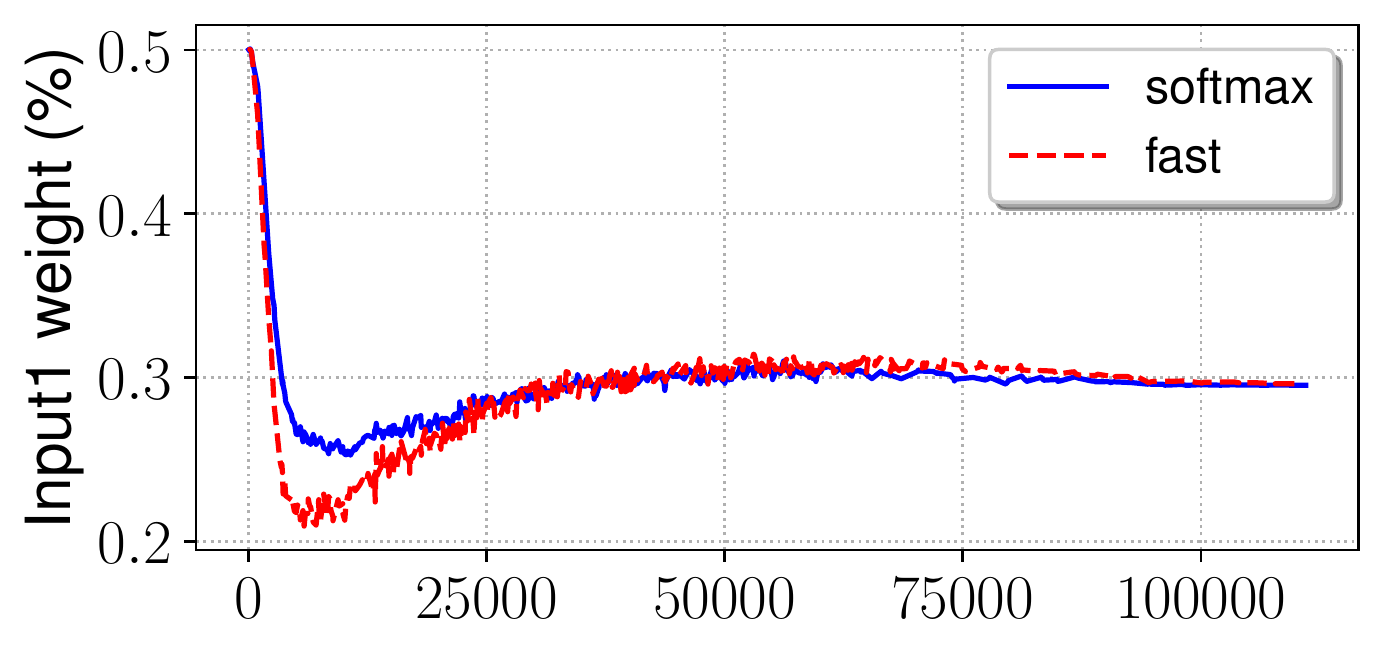}
			\vskip -0.1in
			\caption{Example Node 2}
		\end{subfigure}
        \hfill
        \begin{subfigure}[t]{0.32\textwidth}
			\includegraphics[width=1.0\linewidth]{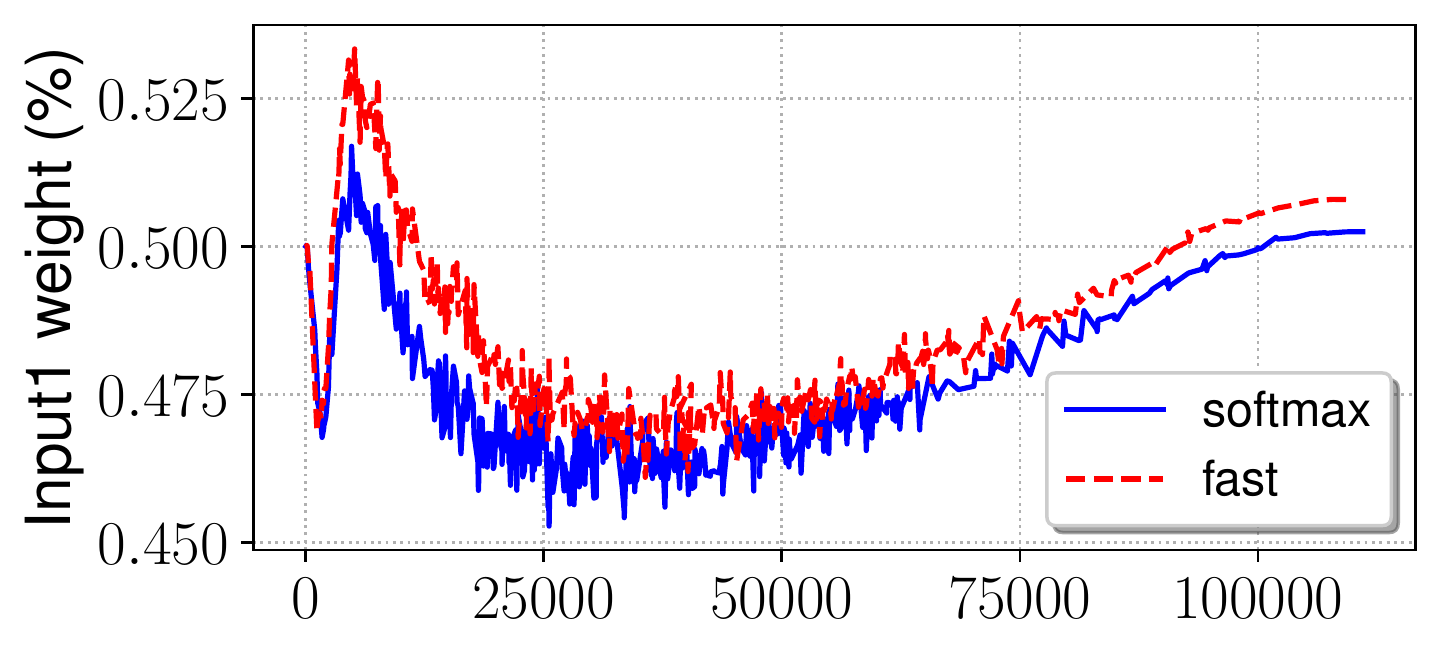}
			\vskip -0.1in
			\caption{Example Node 3}
		\end{subfigure}
	    \vskip -0.1in
        \caption{
        	\BF{Softmax vs. fast normalized feature fusion -- } (a) - (c) shows normalized weights (i.e., importance) during training for three representative nodes; each node has two inputs (input1 \& input2) and their normalized weights always sum up to 1.
        }
        \label{fig:attn}                                                
\end{figure*}

In this section, we ablate various design choices for our proposed EfficientDet. For simplicity, all accuracy results here are for COCO validation set.

\subsection{Disentangling  Backbone and BiFPN}

Since EfficientDet uses both a powerful backbone and a new BiFPN, we want to understand how much each of them contributes to the accuracy and efficiency improvements. Table \ref{tab:backbonefpn} compares the impact of backbone and BiFPN using RetinaNet training settings. Starting from a RetinaNet detector \cite{retinanet17} with ResNet-50 \cite{resnet16} backbone and top-down FPN \cite{fpn17}, we first replace the backbone with EfficientNet-B3, which improves accuracy by about 3 AP with slightly less parameters and FLOPs. By further replacing FPN with our proposed BiFPN, we achieve additional 4 AP gain with much fewer parameters and  FLOPs. These results suggest that EfficientNet backbones and  BiFPN are both crucial for our final models.

\begin{table}[!h]
	\centering
	\resizebox{0.99\columnwidth}{!}{
		\begin{tabular}{l|ccc}
			\toprule[0.15em]
			   &    AP  & Parameters  & FLOPs \\
			\midrule[0.1em]

			ResNet50 + FPN   & 37.0  &  34M		&	97B  \\
			\BF{EfficientNet-B3} + FPN & 	40.3 & 21M & 75B \\
			\BF{EfficientNet-B3} + \BF{BiFPN}  & 44.4 &  12M	&	24B   \\
			\bottomrule[0.15em]
		\end{tabular}
	}
	\caption{
	\textbf{Disentangling backbone and BiFPN -- } Starting from the standard RetinaNet (ResNet50+FPN), we first replace the backbone with EfficientNet-B3, and then replace the baseline FPN with our proposed BiFPN.
   }
	\label{tab:backbonefpn}
	\vskip -0.1in
\end{table}

\subsection{BiFPN Cross-Scale Connections}

Table \ref{tab:bifpncompare} shows the accuracy and model complexity for  feature networks with different cross-scale connections listed in Figure \ref{fig:bifpn}. Notably, the original FPN \cite{fpn17} and PANet \cite{panet18} only have one top-down or bottom-up flow, but for fair comparison, here we repeat each of them multiple times and replace all convs with depthwise separable convs, which is the same as BiFPN. We use the same backbone and class/box prediction network, and the same training settings for all experiments. As we can see, the conventional top-down FPN is inherently limited by the one-way information flow and thus has the lowest accuracy. While repeated FPN+PANet achieves slightly better accuracy than NAS-FPN \cite{nasfpn19}, it also requires more parameters and FLOPs. Our BiFPN achieves similar accuracy as repeated FPN+PANet, but uses much less parameters and FLOPs. With the additional weighted feature fusion, our BiFPN further achieves the best accuracy with fewer parameters and FLOPs.

\begin{table}[!h]                                  
	\centering   	
	\resizebox{0.99\columnwidth}{!}{ 
		\begin{tabular}{l|ccc}                
			\toprule[0.15em]                                                 
			      &    \multirow{2}{*}{AP} &  \#Params  & \#FLOPs  \\
			      &     &  ratio &  ratio \\
			\midrule[0.1em]
			Repeated top-down FPN 	&  42.29	&	1.0x	&	1.0x \\ 
			Repeated FPN+PANet	&	44.08	&	1.0x	&	1.0x	\\
			NAS-FPN	&  43.16   &  0.71x	&	0.72x  \\
			Fully-Connected FPN &   43.06	&	1.24x	&	1.21x   \\
			\bf BiFPN (w/o weighted)	&	\bf 43.94	&	\bf 0.88x	&	\bf 0.67x	\\
			\bf BiFPN (w/ weighted)	&	\bf 44.39	&	\bf 0.88x	&	\bf 0.68x	\\
			\bottomrule[0.15em] 
		\end{tabular}                                                                 
	}
	\caption{                                                                         
	\textbf{Comparison of different feature networks -- } 
	Our weighted BiFPN achieves the best accuracy with fewer parameters  and FLOPs.
   }
	\label{tab:bifpncompare}
	\vskip -0.1in
\end{table}

\subsection{Softmax vs Fast Normalized Fusion}
\label{subsec:ablation-att}
\begin{table}                                                   
	\centering   	
	\resizebox{0.99\columnwidth}{!}{ 
		\begin{tabular}{l|ccc}                
			\toprule[0.15em]                                                 
			\multirow{2}{*}{Model}      &   Softmax Fusion &  Fast Fusion & \multirow{2}{*}{Speedup}  \\
			 & AP  & AP (delta) & \\
			\midrule[0.1em]
			Model1	&  33.96	&	33.85 (-0.11)	&	1.28x \\ 
			Model2	&  43.78   &  43.77 (-0.01)	&	1.26x  \\
			Model3 &   48.79		&	48.74 (-0.05)	&	1.31x   \\
			\bottomrule[0.15em]                                                           
		\end{tabular}                                                                 
	}
	\caption{                                                                         
	\textbf{Comparison of different feature fusion -- }  Our fast fusion achieves similar accuracy as  softmax-based fusion, but runs 28\% - 31\% faster.
   }
	\label{tab:attnacc}
	\vskip -0.1in
\end{table} 
 
As discussed in Section \ref{subsec:attention},  we propose a fast normalized feature fusion approach to get ride of the expensive softmax while retaining the benefits of normalized weights.  Table \ref{tab:attnacc} compares the softmax and fast normalized fusion approaches in three detectors with different model sizes. As shown in the results, our fast normalized fusion approach achieves similar accuracy as the softmax-based fusion, but runs 1.26x - 1.31x faster on GPUs.

In order to further understand the behavior of softmax-based and fast normalized fusion, Figure \ref{fig:attn} illustrates the learned weights for three  feature fusion nodes randomly selected from the BiFPN layers in EfficientDet-D3. Notably, the normalized weights (\eg, $e^{w_i}/\sum_{j}e^{w_j}$ for softmax-based fusion, and $w_i/(\epsilon + \sum_{j}w_j)$ for fast normalized fusion) always sum  up to 1 for all inputs. Interestingly, the normalized weights change rapidly during training, suggesting different features contribute to the feature fusion unequally. Despite the rapid change, our fast normalized fusion approach always shows very similar learning behavior to the softmax-based fusion for all three nodes.

\subsection{Compound Scaling}
\begin{figure}
	    \centering
        \includegraphics[width=0.88\linewidth]{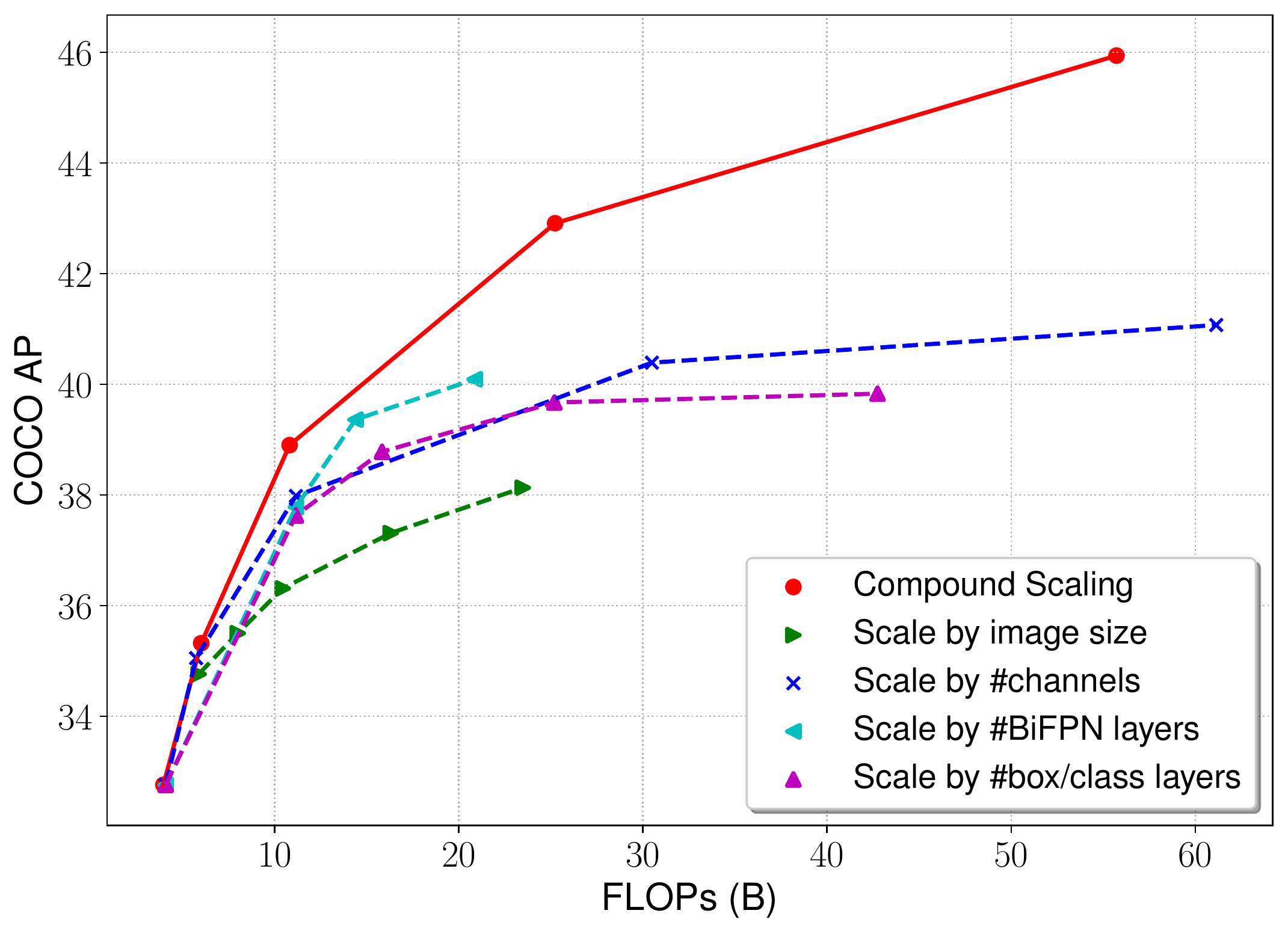}
        \caption{\BF{Comparison of different scaling methods --}
        	 compound scaling achieves better accuracy and efficiency.
         }
        \label{fig:scale-flops}
        \vskip -0.1in
\end{figure}
 
As discussed in section \ref{subsec:scaling}, we employ a compound scaling method to jointly scale up all dimensions of depth/width/resolution for backbone, BiFPN, and box/class prediction networks. Figure \ref{fig:scale-flops} compares our compound scaling with other alternative methods that scale up a single dimension of resolution/depth/width. Although starting from the same baseline detector, our compound scaling method achieves better efficiency than other methods, suggesting the benefits of jointly scaling by better balancing difference architecture dimensions.

\section{Conclusion}
\label{sec:conclude}

In this paper, we systematically study network architecture design choices for efficient object detection, and propose  a weighted  bidirectional feature network and a customized compound scaling method, in order to improve accuracy and efficiency. Based on these optimizations, we develop a new family of detectors, named \emph{EfficientDet}, which consistently achieve better accuracy and efficiency than the prior art across a wide spectrum of resource constraints. In particular, our scaled EfficientDet achieves state-of-the-art accuracy with much fewer parameters and FLOPs than previous object detection and semantic segmentation models. 	%
\section*{Acknowledgements}

Special thanks to Golnaz Ghiasi, Adams Yu, Daiyi Peng  for their help on infrastructure and discussion. We also thank Adam Kraft, Barret Zoph, Ekin D. Cubuk, Hongkun Yu, Jeff Dean, Pengchong Jin, Samy Bengio, Reed Wanderman-Milne, Tsung-Yi Lin, Xianzhi Du, Xiaodan Song, Yunxing Dai, and the Google Brain team. We thank the open source community for the contributions.
 
	{\small
		\bibliographystyle{sty/ieee_fullname}
		\bibliography{cv}

\begin{thebibliography}{10}\itemsep=-1pt

\bibitem{detectron2}
Detectron2.
\newblock {\footnotesize \url{https://github.com/facebookresearch/detectron2}}.
\newblock Accessed: 05/01/2020.

\bibitem{gatelabel17}
Md Amirul~Islam, Mrigank Rochan, Neil~DB Bruce, and Yang Wang.
\newblock Gated feedback refinement network for dense image labeling.
\newblock {\em CVPR}, pages 3751--3759, 2017.

\bibitem{softnms17}
Navaneeth Bodla, Bharat Singh, Rama Chellappa, and Larry~S Davis.
\newblock Soft-nms--improving object detection with one line of code.
\newblock {\em ICCV}, pages 5561--5569, 2017.

\bibitem{multiscaleod16}
Zhaowei Cai, Quanfu Fan, Rogerio~S Feris, and Nuno Vasconcelos.
\newblock A unified multi-scale deep convolutional neural network for fast
  object detection.
\newblock {\em ECCV}, pages 354--370, 2016.

\bibitem{cascade18}
Zhaowei Cai and Nuno Vasconcelos.
\newblock Cascade r-cnn: Delving into high quality object detection.
\newblock {\em CVPR}, pages 6154--6162, 2018.

\bibitem{deeplabv3plus14}
Liang-Chieh Chen, Yukun Zhu, George Papandreou, Florian Schroff, and Hartwig
  Adam.
\newblock Encoder-decoder with atrous separable convolution for semantic image
  segmentation.
\newblock {\em ECCV}, 2018.

\bibitem{xception17}
Fran{\c{c}}ois Chollet.
\newblock Xception: Deep learning with depthwise separable convolutions.
\newblock {\em CVPR}, pages 1610--02357, 2017.

\bibitem{swishsil18}
Stefan Elfwing, Eiji Uchibe, and Kenji Doya.
\newblock Sigmoid-weighted linear units for neural network function
  approximation in reinforcement learning.
\newblock {\em Neural Networks}, 107:3--11, 2018.

\bibitem{pascalvoc12}
Mark Everingham, S.~M.~Ali Eslami, Luc Van~Gool, Christopher K.~I. Williams,
  John Winn, and Andrew Zisserman.
\newblock The pascal visual object classes challenge: A retrospective.
\newblock {\em International Journal of Computer Vision}, 2015.

\bibitem{nasfpn19}
Golnaz Ghiasi, Tsung-Yi Lin, Ruoming Pang, and Quoc~V. Le.
\newblock Nas-fpn: Learning scalable feature pyramid architecture for object
  detection.
\newblock {\em CVPR}, 2019.

\bibitem{fastrcnn15}
Ross Girshick.
\newblock Fast r-cnn.
\newblock {\em ICCV}, 2015.

\bibitem{rethinking19}
Kaiming He, Ross Girshick, and Piotr Doll{\'a}r.
\newblock Rethinking imagenet pre-training.
\newblock {\em ICCV}, 2019.

\bibitem{maskrcnn17}
Kaiming He, Georgia Gkioxari, Piotr Doll{\'a}r, and Ross Girshick.
\newblock Mask r-cnn.
\newblock {\em ICCV}, pages 2980--2988, 2017.

\bibitem{resnet16}
Kaiming He, Xiangyu Zhang, Shaoqing Ren, and Jian Sun.
\newblock Deep residual learning for image recognition.
\newblock {\em CVPR}, pages 770--778, 2016.

\bibitem{gelu16}
Dan Hendrycks and Kevin Gimpel.
\newblock Gaussian error linear units (gelus).
\newblock {\em arXiv preprint arXiv:1606.08415}, 2016.

\bibitem{mobilenetv319}
Andrew Howard, Mark Sandler, Grace Chu, Liang-Chieh Chen, Bo~Chen, Mingxing
  Tan, Weijun Wang, Yukun Zhu, Ruoming Pang, Vijay Vasudevan, Quoc~V. Le, and
  Hartwig Adam.
\newblock Searching for mobilenetv3.
\newblock {\em ICCV}, 2019.

\bibitem{speedod17}
Jonathan Huang, Vivek Rathod, Chen Sun, Menglong Zhu, Anoop Korattikara,
  Alireza Fathi, Ian Fischer, Zbigniew Wojna, Yang Song, Sergio Guadarrama,
  et~al.
\newblock Speed/accuracy trade-offs for modern convolutional object detectors.
\newblock {\em CVPR}, 2017.

\bibitem{parallelFPN18}
Seung-Wook Kim, Hyong-Keun Kook, Jee-Young Sun, Mun-Cheon Kang, and Sung-Jea
  Ko.
\newblock Parallel feature pyramid network for object detection.
\newblock {\em ECCV}, 2018.

\bibitem{panopticfpn19}
Alexander Kirillov, Ross Girshick, Kaiming He, and Piotr Dollár.
\newblock Panoptic feature pyramid networks.
\newblock {\em CVPR}, 2019.

\bibitem{deepfpn18}
Tao Kong, Fuchun Sun, Chuanqi Tan, Huaping Liu, and Wenbing Huang.
\newblock Deep feature pyramid reconfiguration for object detection.
\newblock {\em ECCV}, 2018.

\bibitem{cornetnet18}
Hei Law and Jia Deng.
\newblock Cornernet: Detecting objects as paired keypoints.
\newblock {\em ECCV}, 2018.

\bibitem{pan18}
Hanchao Li, Pengfei Xiong, Jie An, and Lingxue Wang.
\newblock Pyramid attention networks.
\newblock {\em BMVC}, 2018.

\bibitem{fpn17}
Tsung-Yi Lin, Piotr Doll{\'a}r, Ross Girshick, Kaiming He, Bharath Hariharan,
  and Serge Belongie.
\newblock Feature pyramid networks for object detection.
\newblock {\em CVPR}, 2017.

\bibitem{retinanet17}
Tsung-Yi Lin, Piotr Doll{\'a}r, Ross Girshick, Kaiming He, Bharath Hariharan,
  and Serge Belongie.
\newblock Focal loss for dense object detection.
\newblock {\em ICCV}, 2017.

\bibitem{coco14}
Tsung-Yi Lin, Michael Maire, Serge Belongie, James Hays, Pietro Perona, Deva
  Ramanan, Piotr Doll{\'a}r, and C~Lawrence Zitnick.
\newblock Microsoft {COCO}: Common objects in context.
\newblock {\em ECCV}, 2014.

\bibitem{panet18}
Shu Liu, Lu Qi, Haifang Qin, Jianping Shi, and Jiaya Jia.
\newblock Path aggregation network for instance segmentation.
\newblock {\em CVPR}, 2018.

\bibitem{ssd16}
Wei Liu, Dragomir Anguelov, Dumitru Erhan, Christian Szegedy, Scott Reed,
  Cheng-Yang Fu, and Alexander~C Berg.
\newblock {SSD}: Single shot multibox detector.
\newblock {\em ECCV}, 2016.

\bibitem{rethinkprune18}
Zhuang Liu, Mingjie Sun, Tinghui Zhou, Gao Huang, and Trevor Darrell.
\newblock Rethinking the value of network pruning.
\newblock {\em ICLR}, 2019.

\bibitem{yololite18}
Jonathan Pedoeem and Rachel Huang.
\newblock Yolo-lite: a real-time object detection algorithm optimized for
  non-gpu computers.
\newblock {\em arXiv preprint arXiv:1811.05588}, 2018.

\bibitem{megdet18}
Chao Peng, Tete Xiao, Zeming Li, Yuning Jiang, Xiangyu Zhang, Kai Jia, Gang Yu,
  and Jian Sun.
\newblock Megdet: A large mini-batch object detector, 2018.

\bibitem{swish18}
Prajit Ramachandran, Barret Zoph, and Quoc~V Le.
\newblock Searching for activation functions.
\newblock {\em ICLR workshop}, 2018.

\bibitem{amoebanets18}
Esteban Real, Alok Aggarwal, Yanping Huang, and Quoc~V Le.
\newblock Regularized evolution for image classifier architecture search.
\newblock {\em AAAI}, 2019.

\bibitem{yolo17}
Joseph Redmon and Ali Farhadi.
\newblock Yolo9000: better, faster, stronger.
\newblock {\em CVPR}, 2017.

\bibitem{yolov318}
Joseph Redmon and Ali Farhadi.
\newblock Yolov3: An incremental improvement.
\newblock {\em arXiv preprint arXiv:1804.02767}, 2018.

\bibitem{fasterrcnn15}
Shaoqing Ren, Kaiming He, Ross Girshick, and Jian Sun.
\newblock Faster r-cnn: Towards real-time object detection with region proposal
  networks.
\newblock {\em NIPS}, 2015.

\bibitem{overfeat14}
Pierre Sermanet, David Eigen, Xiang Zhang, Micha{\"e}l Mathieu, Rob Fergus, and
  Yann LeCun.
\newblock Overfeat: Integrated recognition, localization and detection using
  convolutional networks.
\newblock {\em ICLR}, 2014.

\bibitem{sepconv14}
Laurent Sifre.
\newblock Rigid-motion scattering for image classification.
\newblock {\em Ph.D. thesis section 6.2}, 2014.

\bibitem{mnas19}
Mingxing Tan, Bo Chen, Ruoming Pang, Vijay Vasudevan, and Quoc~V Le.
\newblock Mnasnet: Platform-aware neural architecture search for mobile.
\newblock {\em CVPR}, 2019.

\bibitem{efficientnet19}
Mingxing Tan and Quoc~V. Le.
\newblock Efficientnet: Rethinking model scaling for convolutional neural
  networks.
\newblock {\em ICML}, 2019.

\bibitem{fcos19}
Zhi Tian, Chunhua Shen, Hao Chen, and Tong He.
\newblock Fcos: Fully convolutional one-stage object detection.
\newblock {\em ICCV}, 2019.

\bibitem{resnext17}
Saining Xie, Ross Girshick, Piotr Doll{\'a}r, Zhuowen Tu, and Kaiming He.
\newblock Aggregated residual transformations for deep neural networks.
\newblock {\em CVPR}, pages 5987--5995, 2017.

\bibitem{m2det19}
Qijie Zhao, Tao Sheng, Yongtao Wang, Zhi Tang, Ying Chen, Ling Cai, and Haibin
  Ling.
\newblock M2det: A single-shot object detector based on multi-level feature
  pyramid network.
\newblock {\em AAAI}, 2019.

\bibitem{stdn18}
Peng Zhou, Bingbing Ni, Cong Geng, Jianguo Hu, and Yi Xu.
\newblock Scale-transferrable object detection.
\newblock {\em CVPR}, pages 528--537, 2018.

\bibitem{objectpoints19}
Xingyi Zhou, Dequan Wang, and Philipp Krähenbühl.
\newblock Objects as points.
\newblock {\em arXiv:1904.07850}, 2019.

\bibitem{odaa19}
Barret Zoph, Ekin~D. Cubuk, Golnaz Ghiasi, Tsung-Yi Lin, Jonathon Shlens, and
  Quoc~V. Le.
\newblock Learning data augmentation strategies for object detection.
\newblock {\em arXiv preprint arXiv:1804.02767}, 2019.

\end{thebibliography}
	}

\section*{Appendix}
\setcounter{section}{1}
\subsection{Hyperparameters}
Neural network architecture and training hyperparamters are both crucial for object detection. Here we ablate two important hyperparamters: training epochs and multi-scale jittering,  using RetinaNet-R50 and our EfficientDet-D1. All other hyperparameters are kept the same as section \ref{sec:results}.

\begin{figure}[h]
    \small
    \centering
    \vskip -0.1in
    \includegraphics[width=0.9\linewidth]{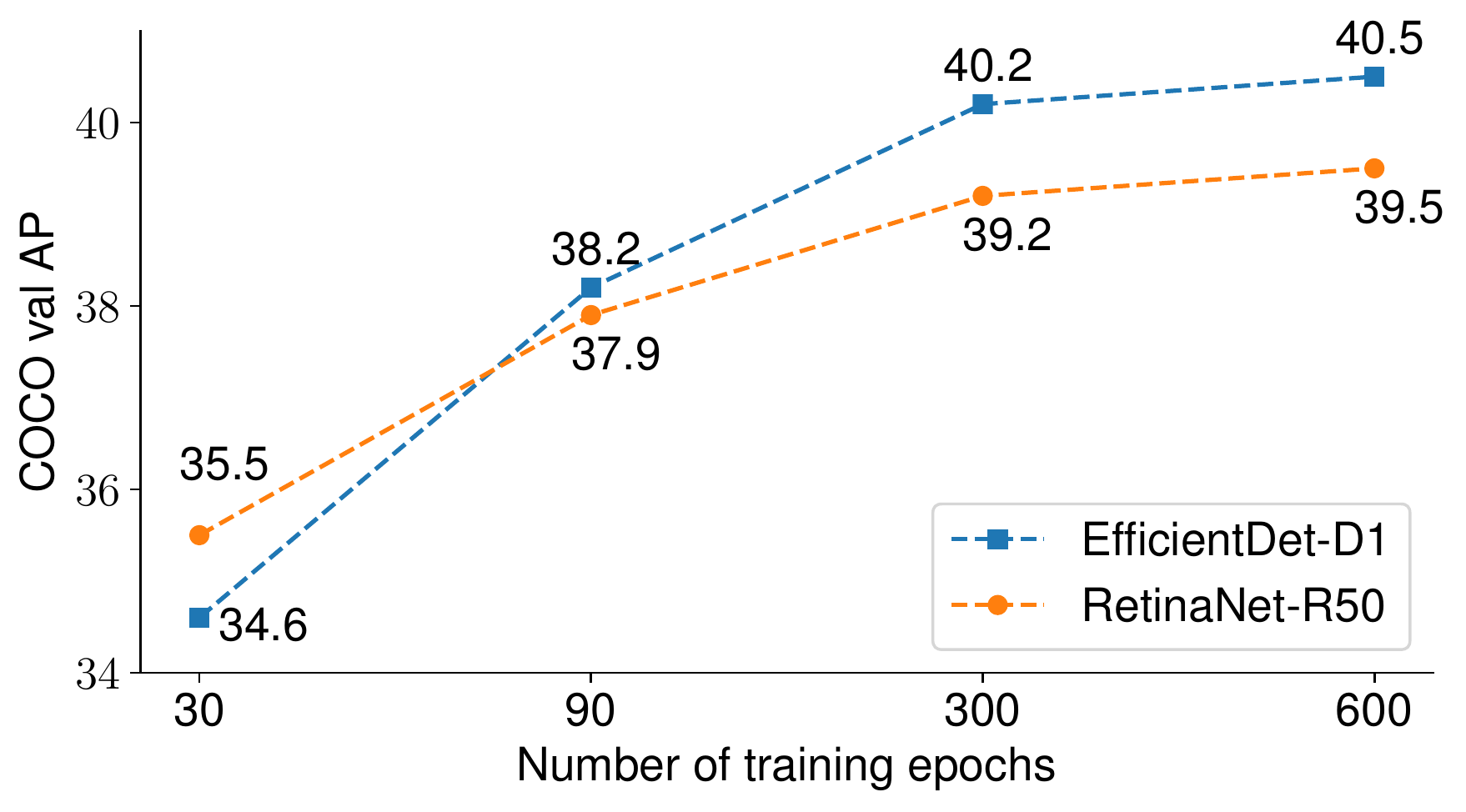}
    \vskip -0.1in
    \caption{\small \BF{Accuracy vs. Training Epochs.} }
    \label{fig:append-epoch}
    \vskip -0.2in
\end{figure}
 
\noindent \paragraph{Training Epochs:} Many previous work only use a small number of epochs: for example, Detectron2 \cite{detectron2} trains each model with 12 epochs (1x schedule) in default, and at most 110 epochs (9x scahedule). Recent work \cite{rethinking19} shows training longer is not helpful if using pretrained backbone networks; however, we observe training longer can significiantly improve accuracy in our settings. Figure \ref{fig:append-epoch} shows the performance comparison for different training epochs. We obseve: (1) both models benefit from longer training until reaching 300 epochs; (2) longer training is particularly important for EfficientDet, perhaps due to its small model size; (3) compared to the default 37 AP \cite{retinanet17}, our reproduced RetinaNet achieves higher accuracy (+2AP) using our training settings. In this paper, we mainly use 300 epochs for the good trade-off between accuracy and training time.

\begin{figure}[h]
    \small
    \centering
    \vskip -0.1in
    \includegraphics[width=0.9\linewidth]{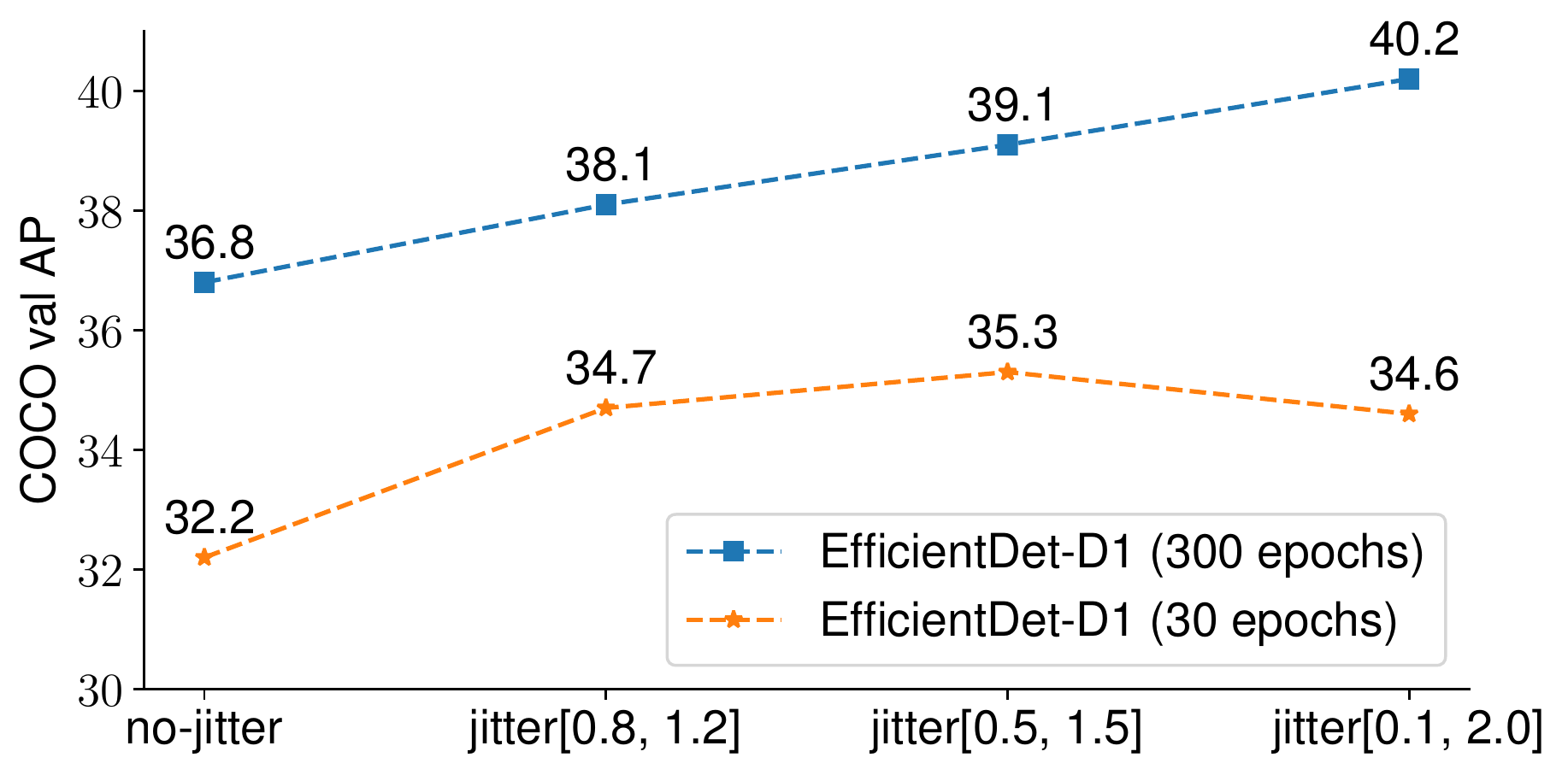}
    \vskip -0.1in
    \caption{\small \BF{Accuracy vs. Scale Jittering.} }
    \label{fig:append-jitter}
    \vskip -0.2in
\end{figure}

\noindent \paragraph{Scale Jittering:} A common training-time augmentation is to first resize images and then crop them into fixed size, known as scale jitterinig. Previous object detectors often use small jitters such as  [0.8, 1.2], which randomly sample a scaling size between 0.8x to 1.2x of the original image size. However, we observe large jitters can improve accuracy if training longer. Figure \ref{fig:append-jitter} shows the results for different jitters: (1) when training with 30 epochs, a small jitter like [0.8, 1.2] performs quite good, and large jitters like [0.1, 2.0] actually hurts accuracy; (2) when training with 300 epochs, large jitters consistently improve accuracy, perhaps due to the stronger regularization. This paper uses a large jitter [0.1, 2.0] for all models.

\subsection{Image Resolutions}
\begin{figure}[h]
    \small
    \centering
    \vskip -0.1in
    \includegraphics[width=0.9\linewidth]{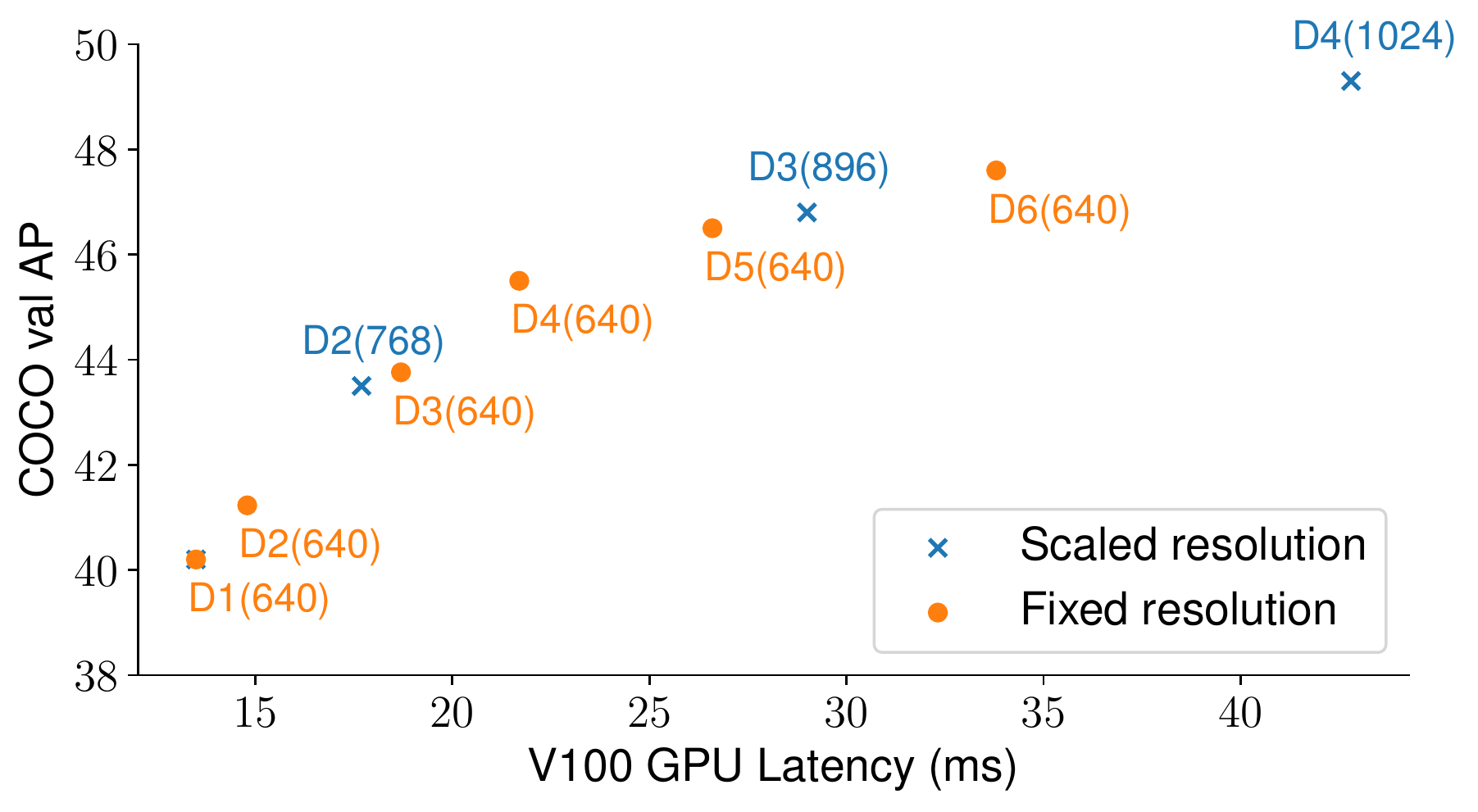}
    \vskip -0.1in
    \caption{\small \BF{Comparison for Fixed and Scaled Resolution} -- \emph{fixed}  denotes 640x640 size and \emph{scaled} denotes increased sizes.}
    \label{fig:append-resolution}
    \vskip -0.2in
\end{figure}

In addition to our compound scaling that progressively increases image sizes, we are also interested in the accuracy-latency trade-offs with fixed image resolutions. Figure \ref{fig:append-resolution} compares EfficientDet-D1 to D6 with fixed and scaled resolutions. Surprisingly, their accuracy-latency trade-offs are very similar even though they have very different preferences: under similar accuracy constraints, models with fixed resolutions require much more parameters, but less activations and peak memory usage, than those with scaled resolutions. With fixed 640x640, our EfficientDet-D6 achieves real-time  47.9AP at 34ms latency.

\end{document}